\documentclass[lettersize,journal]{IEEEtran}
\usepackage{amsmath,amsfonts}
\usepackage{algorithmic}
\usepackage{algorithm}
\usepackage{array}
\usepackage[caption=false,font=normalsize,labelfont=sf,textfont=sf]{subfig}
\usepackage{textcomp}
\usepackage{stfloats}
\usepackage{url}
\usepackage{verbatim}
\usepackage{graphicx}
\usepackage{cite}
\usepackage{float}  
\usepackage{booktabs}
\usepackage[table]{xcolor}
\definecolor{lightblue}{rgb}{0.9, 0.9, 1}
\usepackage{cleveref}
\usepackage{tikz}
\usetikzlibrary{mindmap}
\usepackage{smartdiagram}
\usesmartdiagramlibrary{additions}
\usepackage{forest}
\usetikzlibrary{shadows}
\usepackage{relsize}
\usepackage{color}
\definecolor{lightcoral}{rgb}{0.94, 0.5, 0.5}
\definecolor{lightgreen}{rgb}{0.56, 0.93, 0.56}
\definecolor{lightyellow}{rgb}{0.94, 0.84, 0.6}
\definecolor{brightlavender}{rgb}{0.75, 0.58, 0.89}

\usepackage{subfig}

\begin{document}

\title{Recent Advances in Speech Language Models: A Survey}

\author{Wenqian Cui, Dianzhi Yu, Xiaoqi Jiao, Ziqiao Meng, Guangyan Zhang, Qichao Wang, Yiwen Guo, \\and Irwin King,~\IEEEmembership{Fellow,~IEEE}
\thanks{Wenqian Cui (wenqian.cui@link.cuhk.edu.hk), Dianzhi Yu, and Irwin King (king@cse.cuhk.edu.hk) are with the Department of Computer Science and Engineering, The Chinese University of Hong Kong, Hong Kong, China. 
Xiaoqi Jiao and Guangyan Zhang are with the LIGHTSPEED STUDIOS, Shenzhen, China.
Ziqiao Meng is with the National University of Singapore, Singapore. 
Qichao Wang is with Tencent, Shenzhen, China.
Yiwen Guo is an Independent Researcher.}}


\markboth{Journal of \LaTeX\ Class Files,~Vol.~14, No.~8, August~2021}%
{Shell \MakeLowercase{\textit{et al.}}: A Sample Article Using IEEEtran.cls for IEEE Journals}


\maketitle

\begin{abstract}
Large Language Models (LLMs) have recently garnered significant attention, primarily for their capabilities in text-based interactions. However, natural human interaction often relies on speech, necessitating a shift towards voice-based models. A straightforward approach to achieve this involves a pipeline of ``Automatic Speech Recognition (ASR) + LLM + Text-to-Speech (TTS)", where input speech is transcribed to text, processed by an LLM, and then converted back to speech. Despite being straightforward, this method suffers from inherent limitations, such as information loss during modality conversion, significant latency due to the complex pipeline, and error accumulation across the three stages. To address these issues, Speech Language Models (SpeechLMs)---end-to-end models that generate speech without converting from text---have emerged as a promising alternative. This survey paper provides the first comprehensive overview of recent methodologies for constructing SpeechLMs, detailing the key components of their architecture and the various training recipes integral to their development. Additionally, we systematically survey the various capabilities of SpeechLMs, categorize their evaluation metrics, and discuss the challenges and future research directions in this rapidly evolving field.\footnote{Github: https://github.com/dreamtheater123/Awesome-SpeechLM-Survey}
\end{abstract}

\begin{IEEEkeywords}
Speech Language Models, Speech Interaction, Large Language Models.
\end{IEEEkeywords}

\section{Introduction}
\label{sec:intro}
\begin{figure*}[t]
\centering
\subfloat[Illustration of the ``ASR + LLM + TTS" framework.]{
    \label{fig: architecture of TextLM}
    \includegraphics[scale=0.6,page=3]{figures/SLM_architecture.pdf}
}
\\
\subfloat[Illustration of the architecture of a SpeechLM.]{
    \label{fig: architecture of SLM}    
    \includegraphics[scale=0.6,page=1]{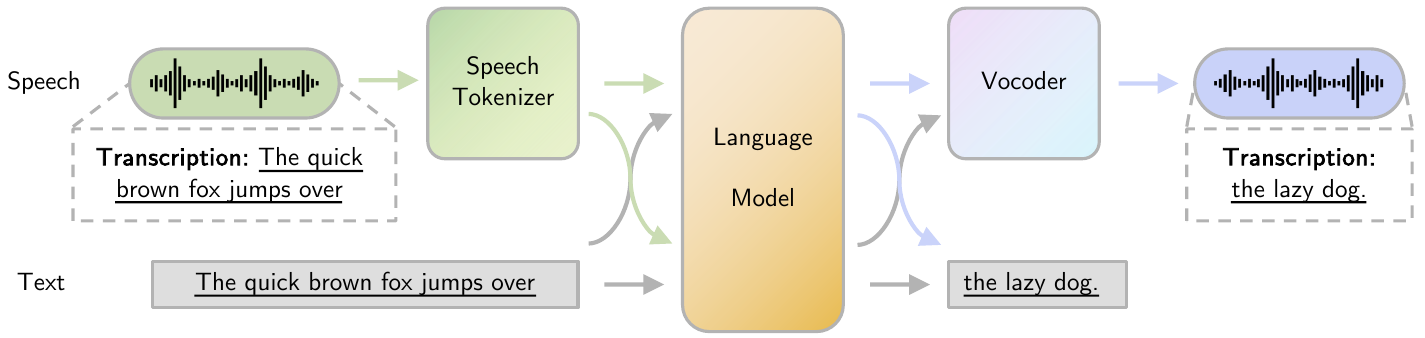}
}
\caption{Architectures of the ``ASR + LLM + TTS" framework and a SpeechLM. We emphasize that, for SpeechLM, the same content can be used across both speech and text modalities, meaning that any input modality can yield any output modality of the same results. This intentional repetition of input/output contents in the figure highlights this point.}
\label{fig:SLM architecture}
\end{figure*}

Large Language Models (LLMs) have demonstrated remarkable capabilities in generating text and performing a wide array of natural language processing tasks \cite{gpt4,llama3,opt}, serving as powerful foundation models for AI-driven language understanding and generation. Their success has also spurred numerous applications in various other domains, yet the reliance solely on text-based modalities presents a significant limitation. This leads to the development of speech-based generative models, which allow to interact with humans more naturally and intuitively. The inclusion of speech not only facilitates real-time voice interactions but also enriches communication by combining both text and speech information \cite{generativedialog2channel,spiritlm}.

Given the extensive mutual information between text and speech, it is natural to modify existing LLMs to enable speech interaction capabilities. A straightforward approach is to adopt an ``Automatic Speech Recognition (ASR) + LLM + Text-to-Speech (TTS)" framework (Figure \ref{fig: architecture of TextLM}) \cite{audiogptASRLLMTTS,hugginggpt}. In this setup, the user's spoken input is first processed by the ASR module, which converts it into text. The LLM then generates a text response based on this transcription. Finally, the TTS module transforms the text response back into speech, which is played back to the user. However, this naive solution mainly suffers from the following three problems.
\begin{enumerate}
    \item \textbf{Information loss.} Speech signals not only contain semantic information (i.e., the meaning of the speech) but also paralinguistic information (e.g., pitch, timbre, tonality, etc.). Putting a text-only LLM in the middle will cause the complete loss of paralinguistic information in the input speech \cite{speechgpt,defossezmoshi}. Recognizing paralinguistic features allows for a more engaging and immersive interaction, as the model can respond with greater context. Moreover, it enables the model to interpret a user’s intent more accurately in certain situations, as the meaning of speech can vary depending on tone.
    \item \textbf{Significant latency.} The sequential operation of ASR, LLM, and TTS leads to considerable delays due to the inherently \textbf{complex structures and pipelines} of these modules \cite{Xie2024MiniOmni,defossezmoshi,llamaomni}. For instance, ASR often includes an additional text generator \cite{Radford2023Robust,dualTransformerASR}, while TTS typically relies on a text tokenizer, both of which increase computational demands. Moreover, implementing advanced decoding techniques, such as beam search for ASR text decoding, can further contribute to delays \cite{Radford2023Robust,dualTransformerASR}. Consequently, developing an end-to-end speech generation model can help significantly minimize latency.
    \item \textbf{Cumulative error.} A staged approach like this can easily lead to cumulative errors throughout the pipeline, particularly in the ASR-LLM stage \cite{audiochatllama,Tang2024SALMONN}. Specifically, transcription errors that occur when converting speech to text in the ASR module can negatively impact the language generation performance of the LLM. Additionally, the TTS performance can deteriorate significantly if the LLM generates text that cannot be synthesized. Consequently, developing a unified architecture plays a crucial role in reducing accumulated errors.
\end{enumerate}

\begin{table}[tb]
    \centering
    \caption{Notations used in this survey.}
    \label{tab:notations}
    \rowcolors{1}{white}{lightblue}
    \begin{tabular}{p{1cm} p{6.5cm}}
        \toprule
        Notation & Description\\
        \midrule
        $\textbf{a}$     & A speech audio waveform\\
        $\hat{\textbf{a}}$ & Model-reconstructed speech audio waveform\\
        $\textbf{t}$     & A text span \\
        $\textbf{M}$     & A multi-modal sequence containing speech and/or text\\
        $\theta$       & Model parameters\\
        $f_E(\cdot)$   & Speech encoder\\
        $d(\cdot)$     & Speech quantizer\\
        $LM$           & Language model\\
        $\textbf{v}$   & Encoded speech representations\\
        $\textbf{s}$   & Speech tokens\\
        $\textbf{m}$   & Multi-modal tokens (speech and/or text)\\
        $V$            & Vocabulary of a language model\\
        $E$            & Embedding matrix of a language model\\
        $\textbf{De}$  & Transformer decoder blocks\\
        $Vo$           & Vocoder\\
        $G$            & Generator in a Generative Adversarial Network\\
        $D$            & Discriminator in a Generative Adversarial Network\\
        $ms$           & Mel-spectrogram of a speech audio waveform\\
        $F_0$          & Fundamental frequency\\
        \bottomrule
    \end{tabular}
\end{table}

The limitations of the naive ASR + LLM + TTS framework have led to the development of Speech Language Models (SpeechLMs, Figure \ref{fig: architecture of SLM}). Unlike the naive framework, SpeechLMs directly encode speech waveforms into tokens or representations, capturing essential features and information from audio (\cref{sec:speechencoder}). Although individual speech tokens may not carry word-level semantic meaning, they capture the semantic information of speech utterances and retain valuable paralinguistic information, which prevents the information loss. SpeechLMs then model these tokens autoregressively, without solely relying on text input, which allows them to use the additional paralinguistic information to generate more expressive and nuanced speech (\cref{sec:languagemodel}). Finally, the generated tokens are synthesized back to speech (\cref{sec:vocoder}). This integrated approach eliminates the need to chain together three separate modules, significantly reducing latency. Additionally, by working directly with the encoded speech tokens, SpeechLMs effectively mitigate the cumulative errors, as their training is integrated with the speech encoding, whereas the training of LLMs (language modeling) is completely independent of the ASR (speech recognition) module in the naive framework.

\begin{figure*}[!t]
\centering
\includegraphics[width=\linewidth,page=5]{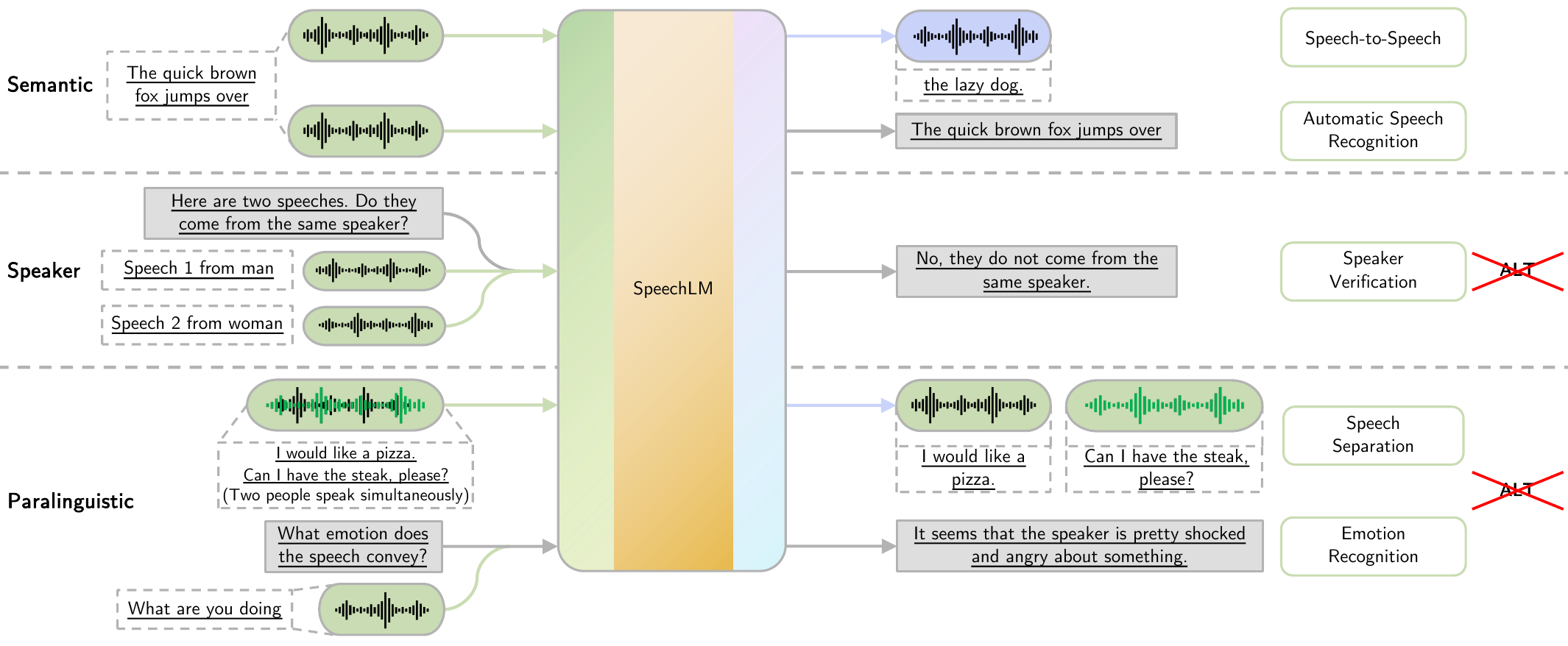}
\caption{Applications of a SpeechLM. We use ALT to represent the ``ASR + LLM + TTS" framework.}
\label{fig:SLM applications}
\end{figure*}

Speech Language Models (SpeechLMs) have the capability to go beyond simple conversational tasks and address more intricate and diverse applications. First, SpeechLMs can capture speaker-specific information and emotional nuances (Figure \ref{fig:SLM applications}), which allows them to distinguish between different speakers during a conversation and to comprehend and generate speech imbued with specific emotional tones. Such advancements are crucial for applications in areas like personalized assistants, emotion-aware systems, and more nuanced human-computer interaction scenarios. 
Second, SpeechLMs can be designed to enable real-time voice interaction, where the model can be interrupted by humans or choose to speak while the user is still speaking, mimicking the dynamics of human conversations more naturally. Furthermore, because SpeechLMs are trained directly on speech data, they have the potential to facilitate communication in rare languages where spoken content is more prevalent than written material.

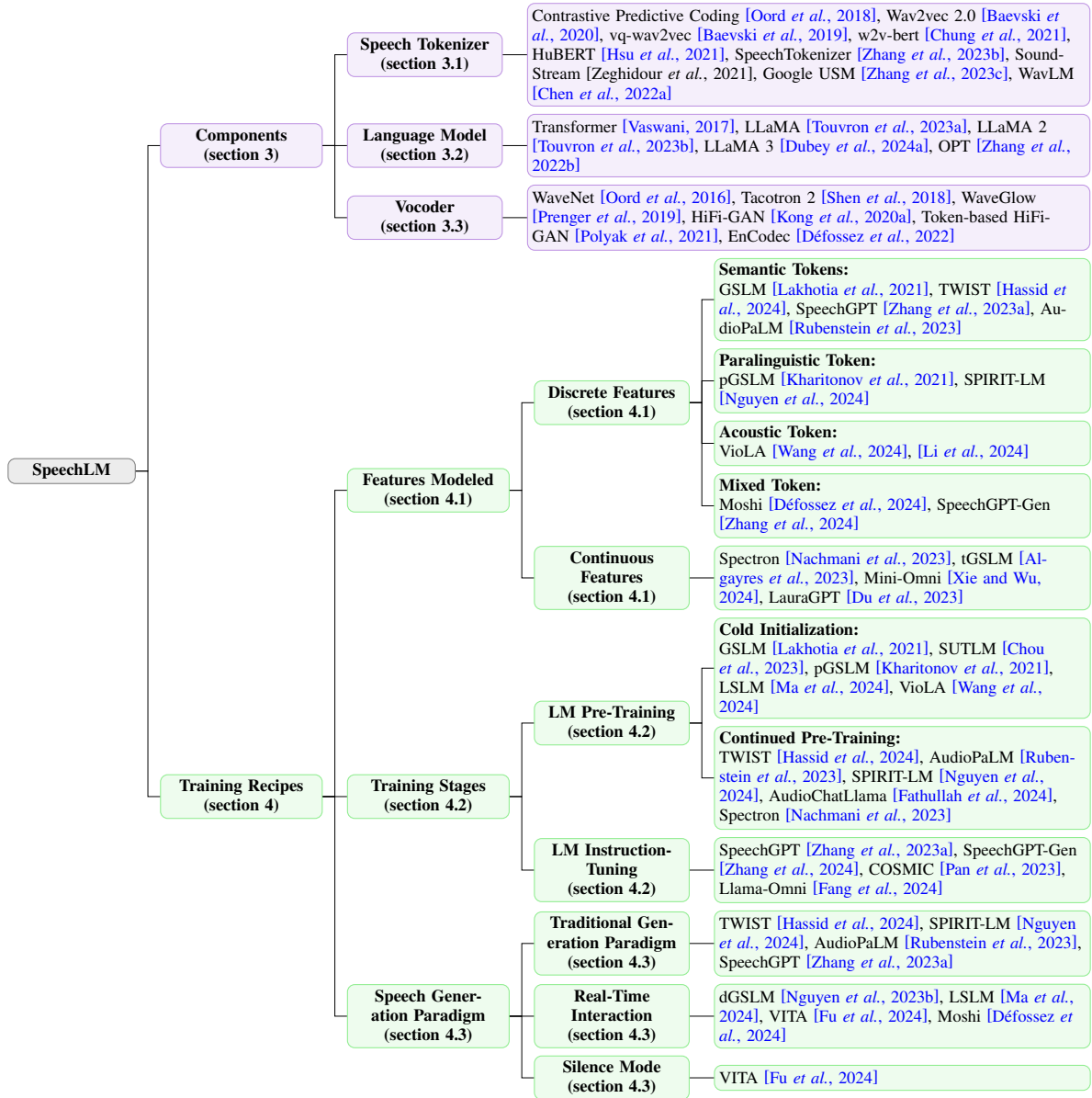
\begin{figure*}[t]
  \centering
  \tikzset{
          my node/.style={
              draw,
              align=center,
              thin,
              text width=1.2cm, 
              rounded corners=3,
          },
          my leaf/.style={
              draw,
              align=left,
              thin,
              text width=8.5cm, 
              rounded corners=3,
          }
  }
  \forestset{
    every leaf node/.style={
      if n children=0{#1}{}
    },
    every tree node/.style={
      if n children=0{minimum width=1em}{#1}
    },
  }
  \begin{forest}
      nonleaf/.style={font=\bfseries\scriptsize},
       for tree={%
          every leaf node={my leaf, font=\scriptsize},
          every tree node={my node, font=\scriptsize, l sep-=4.5pt, l-=1.pt},
          anchor=west,
          inner sep=2pt,
          l sep=10pt, 
          s sep=3pt, 
          fit=tight,
          grow'=east,
          edge={ultra thin},
          parent anchor=east,
          child anchor=west,
          if n children=0{}{nonleaf}, 
          edge path={
              \noexpand\path [draw, \forestoption{edge}] (!u.parent anchor) -- +(5pt,0) |- (.child anchor)\forestoption{edge label};
          },
          if={isodd(n_children())}{
              for children={
                  if={equal(n,(n_children("!u")+1)/2)}{calign with current}{}
              }
          }{}
      }
      [SpeechLM, draw=gray, fill=gray!15, text width=1.7cm, text=black
      [Components \\ ({\cref{sec:components}}), color=brightlavender, fill=brightlavender!15, text width=2.2cm, text=black
            [Speech Tokenizer\\ ({\cref{sec:speechencoder}}), color=brightlavender, fill=brightlavender!15, text width=2.1cm, text=black
            [{
                Contrastive Predictive Coding \color{blue}\cite{cpc}\color{black}, Wav2vec 2.0 \color{blue}\cite{wav2vec2.0}\color{black}, vq-wav2vec \color{blue}\cite{vq-wav2vec}\color{black}, w2v-bert \color{blue}\cite{w2v-bert}\color{black}, HuBERT \color{blue}\cite{hsu2021hubert}\color{black}, SpeechTokenizer \color{blue}\cite{zhang2023speechtokenizer}\color{black}, Mimi \color{blue}\cite{defossezmoshi}\color{black}, SoundStream \cite{soundstream}\color{black}, Google USM \color{blue}\cite{zhang2023googleusm}\color{black}, WavLM \color{blue}\cite{chen2022wavlm}\color{black}
                }, color=brightlavender, fill=brightlavender!15, text width=8cm, text=black]
            ],
            [Language Model\\ ({\cref{sec:languagemodel}}), color=brightlavender, fill=brightlavender!15, text width=2.1cm, text=black
              [{ 
                Transformer \color{blue}\cite{transformer}\color{black}, LLaMA \color{blue}\cite{llama}\color{black}, LLaMA 2 \color{blue}\cite{llama2}\color{black}, LLaMA 3 \color{blue}\cite{llama3}\color{black}, OPT \color{blue}\cite{opt}\color{black}, Qwen2 \color{blue}\cite{yang2024qwen2}\color{black}, GLM \color{blue}\cite{glm2024chatglm}\color{black}, Mixtral \color{blue}\cite{Jiang2024Mixtral}\color{black}
              }, color=brightlavender, fill=brightlavender!15, text width=8cm, text=black],
            ],
            [Vocoder\\ ({\cref{sec:vocoder}}), color=brightlavender, fill=brightlavender!15, text width=2.1cm, text=black
              [{ 
                WaveNet \color{blue}\cite{oord2016wavenet}\color{black}, Tacotron 2 \color{blue}\cite{tacotron2}\color{black}, WaveGlow \color{blue}\cite{prenger2019waveglow}\color{black}, HiFi-GAN \color{blue}\cite{hifigan}\color{black}, Token-based HiFi-GAN \color{blue}\cite{disentangledvocoder}\color{black}, EnCodec \color{blue}\cite{encodec}\color{black}
              }, color=brightlavender, fill=brightlavender!15, text width=8cm, text=black],
            ],
          ]
    [Training Recipes \\ ({\cref{sec:trainingRecipes}}), color=lightgreen, fill=lightgreen!15, text width=2.2cm, text=black
            [Features Modeled \\ ({\cref{sec:featuresmodeled}}), color=lightgreen, fill=lightgreen!15, text width=2.2cm, text=black
            [Discrete Features\\ ({\cref{sec:discretefeatures}}), color=lightgreen, fill=lightgreen!15, text width=2.1cm, text=black
                [{
                    \textbf{Semantic Tokens:} \\GSLM \color{blue}\cite{OnGenerative}\color{black}, TWIST \color{blue}\cite{textuallypretrainSLM}\color{black}, SpeechGPT \color{blue}\cite{speechgpt}\color{black}, AudioPaLM \color{blue}\cite{audiopalm}\color{black}, OmniFlatten \color{blue}\cite{omniflatten}\color{black}, SLAM-Omni \color{blue}\cite{SLAM-Omni}\color{black}, GLM-4-Voice \color{blue}\cite{glm4voice}\color{black}
                    }, color=lightgreen, fill=lightgreen!15, text width=5.3cm, text=black],
                [{
                    \textbf{Paralinguistic Token:} \\pGSLM \color{blue}\cite{prosodyawareSLM}\color{black}, SPIRIT-LM \color{blue}\cite{spiritlm}\color{black}
                    }, color=lightgreen, fill=lightgreen!15, text width=5.3cm, text=black], 
                [{
                    \textbf{Acoustic Token:} \\VioLA \color{blue}\cite{violaCodecLM}\color{black}, Li et al. \color{blue}\cite{codecSLMexamine}\color{black}, Parrot \color{blue}\cite{meng2024parrot_neurips}\color{black}
                    }, color=lightgreen, fill=lightgreen!15, text width=5.3cm, text=black], 
                [{
                    \textbf{Mixed Token:} \\Moshi \color{blue}\cite{defossezmoshi}\color{black}, SpeechGPT-Gen \color{blue}\cite{speechgpt-gen}\color{black}
                    }, color=lightgreen, fill=lightgreen!15, text width=5.3cm, text=black], 
            ],
            [Continuous Features\\ ({\cref{sec:continuousfeatures}}), color=lightgreen, fill=lightgreen!15, text width=2.1cm, text=black
              [{
                Spectron \color{blue}\cite{spectrogramSLM}\color{black}, tGSLM \color{blue}\cite{GSLMcontinuousTokens}\color{black}, Mini-Omni \color{blue}\cite{Xie2024MiniOmni}\color{black}, LauraGPT \color{blue}\cite{Du2023LauraGPT}\color{black}, SLAM-Omni \color{blue}\cite{SLAM-Omni}\color{black}
                }, color=lightgreen, fill=lightgreen!15, text width=5.3cm, text=black],
            ]
          ],
            [Training Stages \\ ({\cref{sec:trainingstages}}), color=lightgreen, fill=lightgreen!15, text width=2.2cm, text=black
            [LM Pre-Training\\ ({\cref{sec:trainingstages}}), color=lightgreen, fill=lightgreen!15, text width=2.1cm, text=black
                [{
                    \textbf{Cold Initialization:} \\GSLM \color{blue}\cite{OnGenerative}\color{black}, SUTLM \color{blue}\cite{jointmodelingSpeechText}\color{black}, pGSLM \color{blue}\cite{prosodyawareSLM}\color{black}, LSLM \color{blue}\cite{LMListenWhileSpeaking}\color{black}, VioLA \color{blue}\cite{violaCodecLM}\color{black}
                    }, color=lightgreen, fill=lightgreen!15, text width=5.3cm, text=black],
                [{
                    \textbf{Continued Pre-Training:} \\TWIST \color{blue}\cite{textuallypretrainSLM}\color{black}, AudioPaLM \color{blue}\cite{audiopalm}\color{black}, SPIRIT-LM \color{blue}\cite{spiritlm}\color{black}, AudioChatLlama \color{blue}\cite{audiochatllama}\color{black}, Spectron \color{blue}\cite{spectrogramSLM}\color{black}, SLAM-Omni \color{blue}\cite{SLAM-Omni}\color{black}, OmniFlatten \color{blue}\cite{omniflatten}\color{black}, Mini-Omni \color{blue}\cite{Xie2024MiniOmni}\color{black}, Mini-Omni 2 \color{blue}\cite{mini-omni2}\color{black}, Freeze-Omni \color{blue}\cite{freeze-omni}\color{black}, Moshi \color{blue}\cite{defossezmoshi}\color{black}, Parrot \color{blue}\cite{meng2024parrot_neurips}\color{black}
                    }, color=lightgreen, fill=lightgreen!15, text width=5.3cm, text=black]
            ],
            [LM Instruction-Tuning\\ ({\cref{sec:trainingstages}}), color=lightgreen, fill=lightgreen!15, text width=2.1cm, text=black
              [{
                SpeechGPT \color{blue}\cite{speechgpt}\color{black}, SpeechGPT-Gen \color{blue}\cite{speechgpt-gen}\color{black}, COSMIC \color{blue}\cite{pan2023cosmicSpeechInstructionTuning}\color{black}, Llama-Omni \color{blue}\cite{llamaomni}\color{black}, Moshi \color{blue}\cite{defossezmoshi}\color{black}
                }, color=lightgreen, fill=lightgreen!15, text width=5.3cm, text=black],
            ]
          ],
          [Speech Generation Paradigm \\ ({\cref{sec:trainingparadigm}}), color=lightgreen, fill=lightgreen!15, text width=2.2cm, text=black
            [Traditional Generation Paradigm\\ ({\cref{sec:trainingparadigm}}), color=lightgreen, fill=lightgreen!15, text width=2.1cm, text=black
              [{ 
                TWIST \color{blue}\cite{textuallypretrainSLM}\color{black}, SPIRIT-LM \color{blue}\cite{spiritlm}\color{black}, AudioPaLM \color{blue}\cite{audiopalm}\color{black}, SpeechGPT \color{blue}\cite{speechgpt}\color{black}
              }, color=lightgreen, fill=lightgreen!15, text width=5.3cm, text=black],
            ],
            [Real-Time Interaction\\ ({\cref{sec:trainingparadigm}}), color=lightgreen, fill=lightgreen!15, text width=2.1cm, text=black
            [{ 
                dGSLM \color{blue}\cite{generativedialog2channel}\color{black}, NTPP \color{blue}\cite{ntpp}\color{black}, LSLM \color{blue}\cite{LMListenWhileSpeaking}\color{black}, VITA \color{blue}\cite{VITAtencentOmniMLLM}\color{black}, Moshi \color{blue}\cite{defossezmoshi}\color{black}, Mini-Omni 2 \color{blue}\cite{mini-omni2}\color{black}, OmniFlatten \color{blue}\cite{omniflatten}\color{black}, SALMONN-omni \color{blue}\cite{salmonn-omni}\color{black}
              }, color=lightgreen, fill=lightgreen!15, text width=5.3cm, text=black],
            ]
            [Interactive Period Recognition\\ ({\cref{sec:trainingparadigm}}), color=lightgreen, fill=lightgreen!15, text width=2.1cm, text=black
            [{ 
                VITA \color{blue}\cite{VITAtencentOmniMLLM}\color{black}, MiniCPM-o 2.6 \color{blue}\cite{MiniCPM-0-2.6}\color{black}, FlexDuo \color{blue}\cite{flexduo-kuaishou}\color{black}
              }, color=lightgreen, fill=lightgreen!15, text width=5.3cm, text=black],
            ]
        ]
      ]
      ]
      \end{forest}
  \caption{Taxonomy of Speech Language Models.}
  \label{fig: Taxonomy}
  \end{figure*}

\textbf{Contributions.} In this survey, we present the first comprehensive overview of recent endeavors in constructing SpeechLMs. We explore the various components that constitute their architecture (Section \ref{sec:components}) and the training recipes (Section \ref{sec:trainingRecipes}) involved in their development. we aim to elucidate the current state of the field by analyzing these models from the above perspectives. Additionally, we survey the downstream applications of SpeechLMs (Section \ref{sec:downstreamApps}), classify metrics to evaluate SpeechLMs (Section \ref{sec:evaluation}), discuss the challenges encountered in this rapidly evolving area, and outline promising future research directions that could drive further advancements in SpeechLM technology (Section \ref{sec:challenges}).
Our contributions are summarized as follows:
\begin{itemize}
    \item We present the first survey in the field of SpeechLMs.
    \item We propose a novel taxonomy (Figure \ref{fig: Taxonomy}) of classifying SpeechLMs from the underlying components and the training recipes.
    \item We propose a novel classification system for the \mbox{evaluation} methods for SpeechLMs.
    \item We identify several challenges in building SpeechLMs.
\end{itemize}

\textbf{Connections with other surveys.} Several surveys have concentrated on traditional speech and audio technologies, such as Spoken Language Understanding (SLU) \cite{SLUsurvey}, audio and speech Self-Supervised Learning (SSL) \cite{audioSSLsurvey,sssl_survey}, and the integration of speech with other modalities \cite{VisuallyGroundedSpeechSurvey,VisualSpeechAnalysisSurvey}. With the rapid advancements in LLMs, some studies review single-modal \cite{LLMsurvey1,LLMsurvey2} and multi-modal \cite{MLLMsurvey1,MLLMsurvey2,MLLMsurveyData} LLMs. Additionally, there are surveys that explore the overlap between the audio modality and LLMs. Latif \textit{et al.} \cite{LargeAudioModelsSurvey} examine LLMs in audio processing, Peng \textit{et al.} \cite{SpeechLLMSurvey_sjtu} review SpeechLLMs within the SLU domain, and Ji \textit{et al.} \cite{wavchatSpokenDialogueModelsSurvey} focus on spoken dialogue systems that encompass speech, sound, and music.

\section{Problem Formulation}

\label{sec:problemformulation}
In this section, we provide a formal definition of Speech Language Models. A \textbf{Speech Language Model (SpeechLM)} is an \textbf{autoregressive foundation model} that processes and generates speech \textbf{end-to-end}, utilizing contextual understanding for coherent sequence generation. This capability allows it to perform a variety of tasks through speech-based interactions. Although SpeechLMs are required to perform end-to-end speech interactions, they can also incorporate text, enabling cross-modal functionalities such as speech-in-text-out and vice versa.
We note that the concept of SpeechLM is in contrast to traditional text-based language models, such as LLM, where the only modality being processed within the model is text. Therefore, to avoid confusion, we call those text-based language models TextLMs throughout this survey.

We offer a unified framework in which SpeechLMs can process and generate speech data, text data, or even interleaved speech and text data. Specifically, a speech audio waveform $\textbf{a} = (a_1, a_2, \ldots, a_Q)$ consists of a sequence of audio samples $a_i \in \mathbb{R}$ of length $Q$, where $1 \leq q \leq Q$. Similarly, a text span $\textbf{t} = (t_1, t_2, \ldots, t_K)$ consists of a sequence of text tokens $t_j$ (word, subword, character, etc.) of length $K$. Let $\textbf{M} = (M_1, M_2, \ldots, M_N)$ denote a multimodal sequence of length $N$, where each element $M_i \in \{a_i, t_j\}$. We define $\textbf{M}^{\text{in}} = (M_1^{\text{in}}, M_2^{\text{in}}, \ldots, M_{N_\text{in}}^{\text{in}})$ as the input multimodal sequence and $\textbf{M}^{\text{out}} = (M_1^{\text{out}}, M_2^{\text{out}}, \ldots, M_{N_\text{out}}^{\text{out}})$ as the output multimodal sequence, where $N_\text{in} \geq 0$ and $N_\text{out} \geq 0$. Then, A SpeechLM parameterized by $\theta$ can then be represented as:
\begin{equation}
\label{eq:SLMdefinition}
\textbf{M}^{\text{out}} = SpeechLM(\textbf{M}^{\text{in}}; \theta).
\end{equation}

\section{Components in SpeechLM}
\label{sec:components}

There are three main components within a SpeechLM, namely speech tokenizer, language model, and token-to-speech synthesizer (vocoder), as illustrated in Figure \ref{fig:SLM architecture}. The fundamental reason for such a three-staged design pattern is to use the language modeling architecture (e.g., decoder-only transformer) to model speech autoregressively in the format of audio waveforms. Since both the input and output of a language model are tokens, additional modules need to be attached to the language model to handle the I/O format. Specifically, the speech tokenizer first transforms continuous audio waveforms into tokens or representations to serve as input to the language model, then the language model performs the next-token prediction based on the input speech tokens. Finally, the vocoder transforms the tokens outputted by the language model back into audio waveforms. We note that our focus here is on how the three components are grouped together to form a SpeechLM rather than a comprehensive overview of each component. Therefore, for speech tokenizer and vocoder, we mainly summarize the methods used in existing SpeechLMs. Table \ref{tab:components_summary} summarizes the popular choices of the three components in various SpeechLM papers.


\subsection{Speech Tokenizer}
\label{sec:speechencoder}

\begin{figure*}[t]
    \centering
    \includegraphics[width=0.99\textwidth]{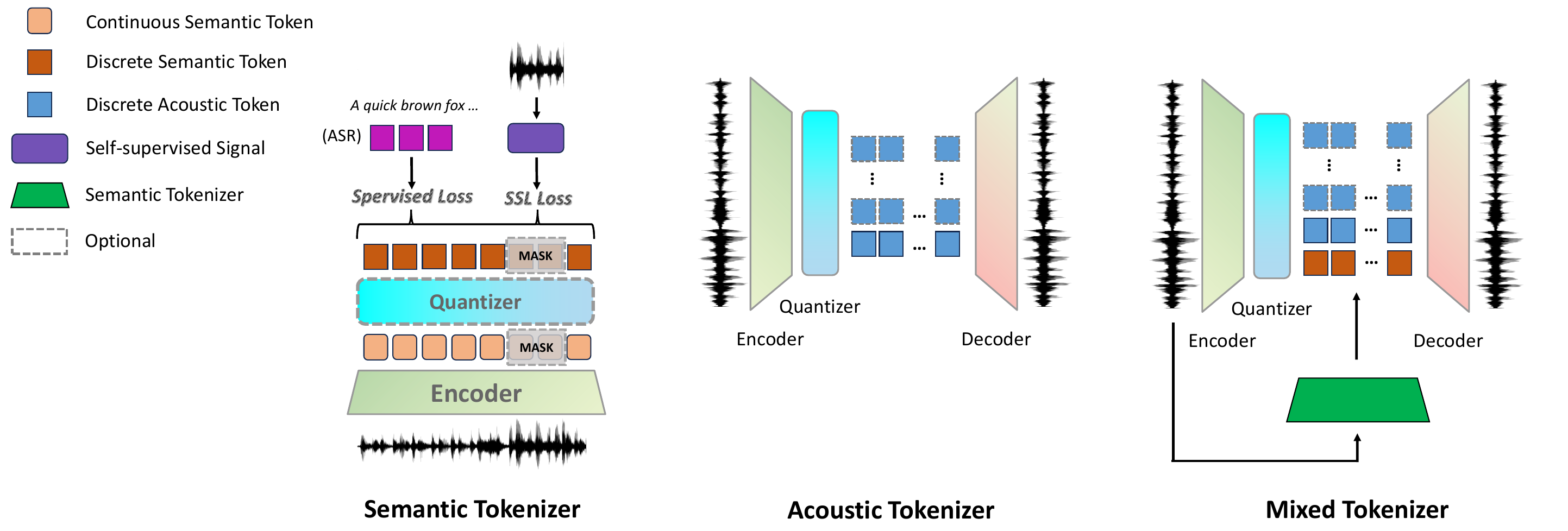} 
    \caption{Illustration of the three types of speech tokenizers.}
    \label{fig:tokenizers}
\end{figure*}

Speech tokenizer is the first component in SpeechLMs, which encodes continuous audio signals (waveforms) into tokens. Speech tokenizer aims to capture essential features of the audio while reducing its dimensionality and allows the audio input to be effectively processed by a language model for autoregressive generation. 
Speech tokenizer operates by encoding the audio segment by segment, producing two possible types of tokens (features): \textbf{discrete tokens} and \textbf{continuous tokens}. Discrete tokens (Section \ref{sec:discretefeatures}) use a specific index to represent each speech segment, while continuous tokens (Section \ref{sec:continuousfeatures}) use an embedding to represent the segment\footnote{For simplicity, we unify both types of representations as "tokens"}. Both token types can be utilized as input for a language model in autoregressive modeling. In this section, we divide speech tokenizers into three categories based on their focus on modeling different aspects of the raw audio. Figure \ref{fig:tokenizers} illustrates the three types of speech tokenizers.

\subsubsection{Semantic Understanding Objective}
\label{sec:semanticUnderstandingTokenizer}
Speech tokenizers designed with a semantic understanding objective aim to convert speech waveforms into tokens that accurately capture the content and meaning of the speech. These tokenizers focus on extracting semantic features from the waveforms, which enhances tasks like ASR.

A semantic understanding speech tokenizer typically comprises a speech encoder and a quantizer. The speech encoder encodes the essential information from the waveform into continuous embeddings. Then, a quantizer is typically incorporated to convert continuous embeddings into discrete indexes. Let $f_E(\cdot)$ denote the speech encoder parameterized by $\theta_{f_E}$, we have $\textbf{v} = f_E(\textbf{a}; \theta_{f_E})$, where $\textbf{v} = (v_1, v_2, \ldots, v_P)$ represents the encoded embeddings. Since $\textbf{v}$ is still continuous, a quantizer $d(\cdot)$ is utilized to discretize the embeddings. Depending on different design choices, the speech tokens \(\textbf{s} = (s_1, s_2, \ldots, s_P)\) can either be derived from $\textbf{a}$ or $\textbf{v}$. Therefore, we have $\textbf{s} = d(\textbf{v}; \theta_d)$ or $\textbf{s} = d(\textbf{a}; \theta_d)$ for discrete tokens and $\textbf{s} = \textbf{v}$ for continuous tokens. After that, $\textbf{s}$ can be used to train the speech tokenizer as a target label (such as masking $\textbf{a}_\text{mask} \subset \textbf{a}$ and reconstructing its corresponding label $\textbf{s}_\text{mask} \subset \textbf{s}$ \cite{hsu2021hubert}) or to train the following language model.

The key design choices lie in how to effectively encode and/or quantize speech into tokens. Wav2vec 2.0 \cite{wav2vec2.0} uses a convolutional encoder followed by a product quantization module \cite{productquantization} to discretize the continuous waveform. Then, a portion of the quantized representations is masked and modeled using a contrastive loss. W2v-BERT \cite{w2v-bert} is built upon wav2vec 2.0 and proposes to use Masked Language Modeling (MLM) loss \cite{bert} in addition to contrastive loss. Similarly, HuBERT \cite{hsu2021hubert} uses the k-means algorithm to cluster the speech utterances into a number of hidden units, and then perform MLM to predict the target hidden units from the masked speech utterances. To better align the representation of text and speech modalities, Google USM \cite{zhang2023googleusm} utilizes text-injection loss \cite{chen2022maestro} at the second pre-training stage to improve the performance and robustness of the downstream tasks.
WavLM \cite{chen2022wavlm} adds the speech denoising objective during pre-training. While the majority of speech tokenizer studies focus on semantic-related tasks such as ASR and TTS, WavLM shows that speech denoising can boost the performance of non-semantic tasks such as speaker verification and speech separation. Section \ref{sec:downstreamApps} presents a full list of downstream tasks.

\subsubsection{Acoustic Generation Objective}
\label{sec:acousticGenerationTokenizer}
Speech tokenizers with an acoustic generation objective focus on capturing the acoustic features necessary for generating high-quality speech waveforms. These tokenizers prioritize the preservation of essential acoustic characteristics over semantic content, making them suitable for speech (re)synthesis tasks.

To generate high-quality speech waveforms, acoustic generation speech tokenizers employ a speech synthesis or speech reconstruction objective. To achieve this, the architecture typically includes an encoder, a quantizer, and a decoder. Same as before, the encoder $f_E(\cdot)$ and quantizer $d(\cdot)$ transform the original waveform into tokens. After that, the decoder $f_D(\cdot)$ reconstructs these tokens back into speech waveforms. This process is represented by $\hat{\textbf{a}} = f_D(\textbf{s}; \theta_{f_E})$, where $\hat{\textbf{a}}$ is the generated or reconstructed waveform.

Neural audio codecs are very suitable for and are primarily employed as acoustic generation speech tokenizers \cite{soundstream,encodec}. These codecs utilize the advanced modeling capabilities of deep neural networks to compress audio waveforms into a compact representation, typically in the form of discrete tokens. Using the encoder-quantizer-decoder architecture, the encoder compresses the audio into latent representations, the quantizer discretizes these representations (commonly through vector quantization (VQ) \cite{vq-vae} or residual vector quantization (RVQ) \cite{soundstream}), and the decoder reconstructs the discrete tokens back into audio waveforms. Therefore, the encoder and/or the quantizer are utilized as an acoustic speech tokenizer.

\subsubsection{Mixed Objective}
\label{sec:mixedTokenizer}
Speech tokenizers with a mixed objective aim to balance both semantic understanding and acoustic generation. The goal is to harness the advantages of both types of tokenizers. Currently, the development of these tokenizers is in its early stages. Most existing mixed speech tokenizers primarily adopt the architecture of acoustic generation speech tokenizers and focus on distilling information from semantic tokenizers into the acoustic tokenizer. SpeechTokenizer \cite{zhang2023speechtokenizer} utilizes the RVQ-GAN \cite{encodec,soundstream} architecture, distilling semantic information from HuBERT \cite{hsu2021hubert} to the first layer of RVQ. Inspired by SpeechTokenizer, Mimi \cite{defossezmoshi} employs a single VQ to extract information from WavLM \cite{chen2022wavlm} and incorporates another RVQ module to learn the acoustic information.

In the following paragraphs, we present the notations for three representative speech tokenizers, each exemplifying a distinct category. We examine HuBERT as a semantic objective tokenizer, Encodec as an acoustic objective tokenizer, and SpeechTokenizer as a mixed objective tokenizer.

\textbf{HuBERT.} As a representative semantic objective tokenizer, HuBERT \cite{hsu2021hubert} employs a feature encoder \( f_E \) to transform raw audio waveforms \( \mathbf{a} \) into continuous embeddings \( \mathbf{v} \), i.e., \( f_E(\mathbf{a}; \theta_{f_E}) = \mathbf{v} \). These embeddings are then quantized into discrete speech tokens \( \mathbf{s} \) via k-means clustering of MFCC features, denoted as \( d(\mathrm{MFCC}(\mathbf{a}); \theta_d) = \mathbf{s} \). The model is trained with a masked prediction objective, which seeks to maximize the likelihood of the correct token at masked positions: 
\begin{equation}
    \mathcal{L}(\theta) = \mathbb{E}_{\mathbf{a} \sim \mathcal{D}} \left[ \sum_{i \in \mathcal{M}} -\log p(s_i \mid \mathbf{v}_{\setminus\mathcal{M}}; \theta) \right],
\end{equation}
where \( \mathcal{M} \) denotes the masked indices. HuBERT further refines its speech tokens iteratively, updating the encoder and discretizer parameters at each step as 
\begin{equation}
\mathbf{s}^{(n+1)} = d(f_E(\mathbf{a}; \theta_{f_E}^{(n)}); \theta_d^{(n)}).
\end{equation}
This iterative process enables the learning of increasingly meaningful speech representations.

\textbf{Encodec.} As a representative acoustic objective tokenizer, EnCodec \cite{encodec} employs a convolutional encoder-decoder architecture with residual vector quantization (RVQ). The encoder \( f_E \) maps the raw audio waveform \( \mathbf{a} \) to continuous embeddings \( \mathbf{v} \), i.e., \( \mathbf{v} = f_E(\mathbf{a}; \theta_{f_E}) \). These embeddings are then discretized using a multi-stage RVQ, where each stage \( r \) quantizes the residual from the previous stage: 
\begin{align}
\mathbf{s} = d(\mathbf{v}; \theta_d) = \big(&d_1(\mathbf{v}; \theta_{d_1}),\ d_2(\mathbf{v} - \hat{\mathbf{v}}_1; \theta_{d_2}), \notag \\
&\ldots,\ d_R(\mathbf{v} - \sum_{r=1}^{R-1}\hat{\mathbf{v}}_r; \theta_{d_R})\big),
\end{align}
with \( \hat{\mathbf{v}}_r \) denoting the quantized embedding at stage \( r \). The decoder \( f_D \) reconstructs the audio waveform from the quantized tokens, \( \hat{\mathbf{a}} = f_D(\mathbf{s}; \theta_{f_D}) \). This design enables EnCodec to produce discrete acoustic tokens that retain high-fidelity audio information suitable for downstream modeling.

\textbf{SpeechTokenizer.} 
As a representative mixed objective tokenizer, SpeechTokenizer \cite{zhang2023speechtokenizer} combines semantic and acoustic objectives by leveraging both HuBERT and residual vector quantization (RVQ) mechanisms. The encoder \( f_E \) first transforms the input audio waveform \(\mathbf{a}\) into continuous embeddings \( \mathbf{v}\), i.e., \( \mathbf{v} = f_E(\mathbf{a}; \theta_{f_E}) \). Discretization is performed via a multi-stage RVQ. The discretization process uses a multi-stage RVQ, which operates similarly to Encodec, except that the first RVQ stage distills tokens derived from HuBERT, while the subsequent stages quantize the residuals. This hybrid approach enables SpeechTokenizer to capture both high-level semantic and low-level acoustic information for robust speech representation learning.

\begin{table*}[t]
    \centering
    \rowcolors{1}{white}{lightblue}
    \caption{Summarization of the architectural choice of speech tokenizer, language model, and vocoder in popular SpeechLMs. ``-'' represents non-existence or not indicated, * means the architecture is mainly based on the written one, ``A, B" means the authors experimented with both ``A" and ``B" as the component, and ``A + B" means ``A" and ``B" are combined to serve as the component.}
\scalebox{0.84}{
    \begin{tabular}{lp{6cm}p{5cm}p{6cm}}
        \toprule
        Approach  & Speech Tokenizer & Language Model & Vocoder \\
        \midrule
        Kimi-Audio \cite{kimiaudio}     &  Whisper Encoder \cite{Radford2023Robust}  + Linear Projector     &    Qwen2.5 \cite{qwen2.5TechnicalReport}   & BigVGAN \cite{lee2022bigvgan}    \\
        Qwen2.5-Omni \cite{qwen2.5omni}     &  Whisper     &    Qwen2.5   & Talker + Codec Decoder \cite{qwen2.5omni}    \\
        Minmo \cite{minmo}     &  SenseVoice \cite{SpeechTeam2024FunAudioLLM}     &    Qwen2.5   & CosyVoice 2 \cite{cosyvoice2}    \\
        Lyra \cite{lyra}     &  Whisper~\cite{Radford2023Robust}     &    Qwen2-VL~\cite{qwen2vl}   & HuBERT + HiFi-GAN    \\
        Flow-Omni \cite{flow-Omni}     &  Whisper Encoder  + Linear Projector     &    Qwen2~\cite{yang2024qwen2}   & Flow Matching (Transformer + MLP) + HiFi-GAN    \\
        SLAM-Omni \cite{SLAM-Omni}     &  Whisper Encoder + Linear Projector     &    Qwen2~\cite{yang2024qwen2}   & -    \\
        OmniFlatten \cite{omniflatten}     &  CosyVoice Encoder~\cite{du2024cosyvoice}    &    Qwen2   & CosyVoice Decoder \cite{du2024cosyvoice}    \\
        SyncLLM \cite{SyncLLM}     &  HuBERT \cite{hsu2021hubert}     &    LLaMA-3~\cite{llama3}   & HiFi-GAN \cite{hifigan,disentangledvocoder}   \\
        EMOVA \cite{emova_huawei}     &  SPIRAL~\cite{huang2022spiral}    &   LLaMA-3    &   VITS~\cite{kim2021conditional}\\
        Freeze-Omni \cite{freeze-omni}     &  Transformer \cite{transformer}    &  Qwen2     &  TiCodec~\cite{ren2024fewer} \\
        IntrinsicVoice \cite{zhang2024intrinsicvoice}     &  HuBERT    &  Qwen2     &  HiFi-GAN \\
        Mini-Omni2 \cite{mini-omni2}     &   Whisper   &    Qwen2   &  Mini-Omni \cite{Xie2024MiniOmni}\\
        SALMONN-omni \cite{salmonn-omni}     & Mamba Streaming Encoder \cite{mamba}     & -      & VoiceCraft \cite{voicecraft} + Codec Decoder  \\
        Zeng et al. \cite{zhipuScalingInterleaving}     & Whisper + VQ    &   GLM~\cite{glm2024chatglm}    & CosyVoice  \\
        NTPP \cite{ntpp}     &  VQ-VAE    &  LLaMA-3, Mistral, Gemma 2     & HiFi-GAN  \\
        GPST \cite{hierarchical_transformer}     &  EnCodec \cite{encodec}     &    Transformer   & Codec Decoder    \\
        GLM-4-Voice \cite{glm4voice}  &   Whisper + VQ \cite{defossezmoshi}   &   GLM-4-9B-Base \cite{glm2024chatglm}   &   CosyVoice  \\
        Moshi \cite{defossezmoshi}  &   Mimi \cite{defossezmoshi}   &   Transformer*   &   Mimi  \\
        VITA \cite{VITAtencentOmniMLLM}  &    CNN + Transformer + MLP \cite{VITAtencentOmniMLLM}    &   Mixtral~\cite{Jiang2024Mixtral}   &  Text-to-Speech Toolkit \cite{VITAtencentOmniMLLM}   \\
        LSLM \cite{LMListenWhileSpeaking} & vq-wav2vec \cite{vq-wav2vec}          & Decoder-Only Transformer        & UniVATS \cite{du2024unicats} \\
        S\textsc{pi}R\textsc{it}-LM \cite{spiritlm}     & HuBERT, VQ-VAE \cite{vq-vae}, speechprop          & LLaMA-2 \cite{llama2}         & HiFi-GAN \\
        TWIST \cite{textuallypretrainSLM}    & HuBERT          & OPT \cite{opt}, LLaMA \cite{llama}        & HiFi-GAN \\
        PSLM \cite{pslm}    & HuBERT          & NekoMata \cite{nekomata}        & HiFi-GAN \\
        VOXTLM \cite{maiti2024voxtlm}     & HuBERT          & OPT \cite{opt}        & HiFi-GAN \\
        Voicebox   \cite{Le2024Voicebox}  &     EnCodec   &   Transformer* \cite{transformer}   &  HiFi-GAN   \\
        Park et al. \cite{park2024let}  &     AV-HuBERT \cite{avhubert}     &   OPT   &   HiFi-GAN  \\
        USDM \cite{USDMparalinguistics}  &     XLS-R \cite{xls-r}     &   Mistral   &   Voicebox \cite{voicebox}  \\
        VioLA \cite{violaCodecLM}  &     EnCodec     &   Transformer*   &   Codec Decoder \cite{encodec}  \\
        FunAudioLLM  \cite{SpeechTeam2024FunAudioLLM}  &   SAN-M~\cite{gao2020san}     &  Transformer*    &  HiFTNet~\cite{li2023hiftnet}   \\
        SpeechGPT-Gen  \cite{speechgpt-gen}  &  SpeechTokenizer \cite{zhang2023speechtokenizer}      &   LLaMA-2     &  SpeechTokenizer decoder \cite{zhang2023speechtokenizer}   \\
        ICoT \cite{icot}  & SpeechTokenizer          & LLaMA-2        & SoundStorm \\
        AnyGPT \cite{anygpt}  & SpeechTokenizer          & LLaMA-2        & SoundStorm \\
        LauraGPT \cite{Du2023LauraGPT}  &   Conformer*      &  Qwen~\cite{bai2023qwen}    &  Transformer + Codec Decoder    \\
        Spectron  \cite{spectrogramSLM}  &    Conformer*     &    PaLM 2*~\cite{anil2023palm}   &  WaveFit~\cite{koizumi2023wavefit}   \\
        AudioLM \cite{audiolm}  &  w2v-BERT \cite{w2v-bert}      &   Decoder-Only Transformer*   &  SoundStream* \cite{soundstream}  \\
        UniAudio   \cite{Yang2023Uniaudio}  &  EnCodec, Hifi-codec~\cite{yang2023hifi}, Improved RVQGAN~\cite{kumar2023highfidelity}      &  Transformer*    &    Codec Decoder \\
        Llama-Omni \cite{llamaomni}  &  Whisper      &  LLaMA-3.1    &  HiFi-GAN   \\
        Mini-Omni   \cite{Xie2024MiniOmni}  &    Whisper + ASR Adapter \cite{Xie2024MiniOmni}     &  Qwen2    &   TTS Adapter \cite{Xie2024MiniOmni}  \\
        tGSLM \cite{GSLMcontinuousTokens}  & Segmentation + SSE \cite{SSE} + Lexical embedder          & Transformer*        & Tacotron-2 + Waveglow \cite{tacotron2,prenger2019waveglow} \\
        SpeechGPT \cite{speechgpt}  & HuBERT          & LLaMA        & HiFi-GAN \\
        dGSLM \cite{generativedialog2channel} & HuBERT          & Dialogue Transformer \cite{generativedialog2channel}        & HiFi-GAN \\
        SUTLM    \cite{jointmodelingSpeechText}  &    HuBERT    &   Transformer*   &   -  \\
        pGSLM \cite{prosodyawareSLM}     & HuBERT         & MS-TLM \cite{prosodyawareSLM}        & HiFi-GAN \\
        GSLM \cite{OnGenerative}     & HuBERT, CPC \cite{cpc}, Wav2vec 2.0 \cite{wav2vec2.0}           & Transformer*        & Tacotron-2 + Waveglow \\
        \bottomrule
    \end{tabular}
}
    \label{tab:components_summary}
\end{table*}

\subsection{Language Model}
\label{sec:languagemodel}
Due to the success of TextLMs \cite{gpt4,gemini,llama3}, most SpeechLMs follow their architectures. They primarily employ transformers \cite{transformer} or decoder-only architectures (such as OPT \cite{opt}, LLaMA \cite{llama}) to generate speech in an autoregressive manner. 
To formally define it, given $|V_t|$ as the vocabulary size and $h$ as the hidden dimension, a typical text-based decoder-only transformer language model consists of an embedding matrix $E_t \in \mathbb{R}^{|V_t| \times h}$, a sequence of \( L \) transformer decoder blocks \(\textbf{De} = \{ De_1, De_2, \ldots, De_L \} \), and an output embedding matrix $E'_t \in \mathbb{R}^{h \times |V_t|}$. Therefore, the language model (LM) can be represented as 
\begin{equation}
\textbf{t}^{\text{out}} \sim \text{LM}(\textbf{t}^{\text{in}}, (E_t, \textbf{De}, E'_t)).
\end{equation}

To adapt the language model to generate speech, the original text tokenizer is changed to the speech tokenizers illustrated in section \ref{sec:speechencoder}. When using \textbf{discrete tokens}, $E_t \in \mathbb{R}^{|V_t| \times h}$ is changed to a speech embedding matrix $E_s \in \mathbb{R}^{|V_s| \times h}$, where $|V_s|$ represents the vocabulary size of the speech tokenizer. The output embedding matrix is also changed from $E'_t \in \mathbb{R}^{h \times |V_t|}$ to $E'_s \in \mathbb{R}^{h \times |V_s|}$. As a result, the language model in a SpeechLM is represented as 
\begin{equation}
\textbf{s}^{\text{out}} \sim \text{LM}(\textbf{s}^{\text{in}}, (E_s, \textbf{De}, E'_s)).
\end{equation}

Because the language model architecture of SpeechLMs is borrowed from TextLMs, it is natural that the resulting model can jointly model both text and speech modalities \cite{spiritlm,speechgpt}. To achieve this, a naive and most adopted approach is to expand the vocabulary of the original TextLM to incorporate both text and speech tokens. Specifically, the speech embedding matrix is usually appended to the end of the text embedding matrix, resulting in a larger embedding matrix $E_m \in \mathbb{R}^{(|V_t|+|V_s|) \times h}$. Let $\textbf{m}$ be a token sequence containing both speech and text tokens, the resulting language model becomes
\begin{equation}
\textbf{m}^{\text{out}} \sim \text{LM}(\textbf{m}^{\text{in}}, (E_j, \textbf{De}, E'_j)).
\end{equation}
By doing so, the model can generate both text and speech in a single sequence, enabling much more diverse applications (see \cref{sec:downstreamApps}).
In contrast, when modeling with \textbf{continuous tokens}, the embeddings derived from the speech tokenizer are directly fed into the language model. In this case, the architecture of the language model remains unchanged.



\subsection{Token-to-Speech Synthesizer (Vocoder)}
\label{sec:vocoder}
After the tokens have been autoregressively generated by the language model component, a token-to-speech module, often known as vocoder, is utilized to synthesize all the speech tokens back into speech waveforms. This process involves converting the linguistic and paralinguistic information represented by the generated speech tokens into audio waveforms that can be heard. This can be seen as a reverse process to the speech tokenizer and therefore can be represented as
\begin{equation}
\textbf{a} = Vo(\textbf{s};\theta_{V_o}), 
\end{equation}
where $Vo$ is the vocoder model parameterized by $\theta_{V_o}$. 

The pipeline of the SpeechLM vocoder can vary depending on the underlying vocoder model. There are two main pipelines: Direct synthesis and input-enhanced synthesis. \textbf{Direct synthesis} is the pipeline where the vocoder directly converts speech tokens generated by the language model into audio waveforms. For example, Polyak \textit{et al.} \cite{disentangledvocoder} adapts the HiFi-GAN \cite{hifigan} architecture and takes speech tokens as inputs. In contrast, \textbf{input-enhanced synthesis} employs an additional module to transform the tokens into a continuous latent representation before they are fed into the vocoder \cite{seedTTS,tortoiseTTS}. The main reason for using this pipeline is that vocoders typically require intermediate audio representations, such as mel-spectrograms \cite{kumar2019melgan,hifigan,lee2022bigvgan}, as input. For example, CosyVoice \cite{du2024cosyvoice} introduces a Conditional Flow-Matching (CFM) model to convert speech tokens into mel-spectrogram, and then leverages a HiFi-GAN to synthesize the final waveform. When comparing the two pipelines, direct synthesis is generally simpler and faster than input-enhanced synthesis. However, the choice of pipeline depends on the type of tokens used as input. Tokens from acoustic generation tokenizers contain sufficient acoustic information, making them suitable for direct synthesis. Conversely, tokens from semantic understanding tokenizers provide rich semantic information but lack fine acoustic details, particularly in higher frequencies. Therefore, these tokens are better enhanced into an acoustic-rich representation, such as mel-spectrograms, before synthesizing the final speech.

Vocoders can be categorized by their architectural choice. In the following sections, we summarize vocoders that are mostly adopted in the development of SpeechLMs.

\subsubsection{GAN-based Vocoder}
Generative Adversarial Network (GAN) is the most adopted architecture of the vocoders \cite{kumar2019melgan,hifigan,disentangledvocoder,fre-gan,lee2022bigvgan}. It is well known for its fast and high-fidelity generation in speech synthesis tasks. The architecture of GAN includes a generator and a discriminator. Specifically, the generator creates realistic audio waveforms from random noise or input features, while the discriminator evaluates the authenticity of the generated audio against real audio samples. 

To utilize GAN to synthesize high-fidelity speech, various training objectives are designed, focusing on different aspects. First, \textbf{GAN loss} is utilized as the fundamental objective for the operation of the generator and the discriminator. Specifically, the typical choice of GAN loss for the generator ($G$) and discriminator ($D$) is to use the least squares loss function. The GAN loss for the generator ($\mathcal{L}_{\text{GAN}}(G; D)$) and the discriminator ($\mathcal{L}_{\text{GAN}}(D; G)$) are
\begin{equation}
\label{eq:ganlossGenerator}
    \mathcal{L}_{\text{GAN}}(G; D) = \mathbb{E}_{ms} \left[ \left( D(G(ms)) - 1 \right)^2 \right]
\end{equation}
and
\begin{equation}
\label{eq:ganlossDiscriminator}
    \mathcal{L}_{\text{GAN}}(D; G) = \mathbb{E}_{(x, ms)} \left[ \left( D(x) - 1 \right)^2 + \left( D(G(ms)) \right)^2 \right], 
\end{equation}
respectively. In these loss functions, $x$ represents the ground truth audio and $ms$ represents its mel-spectrogram. Second, most GAN-based vocoders synthesize speech waveform from mel-spectrograms, so \textbf{mel-spectrogram loss} is proposed to align the mel-spectrogram synthesized by the generator and the mel-spectrogram transformed from the ground-truth waveform, in order to improve the fidelity of the generated speech. Mel-spectrogram loss ($\mathcal{L}_{\text{Mel}}(G)$) works by minimizing the L1 distance between the two versions of mel-spectrograms mentioned above. Its formula is shown below:
\begin{equation}
\label{eq:mel-specloss}
    \mathcal{L}_{\text{Mel}}(G) = \mathbb{E}_{(x, ms)} \left[ \left\| \phi(x) - \phi(G(ms)) \right\|_1 \right], 
\end{equation}
where $\phi(\cdot)$ is the function to transform a waveform into the corresponding mel-spectrogram. Third, to further enhance the generation fidelity, \textbf{feature matching loss} ($\mathcal{L}_{FM}(G;D)$) is proposed to align the discriminator-encoded features of the ground truth sample and the generated sample with L1 distance, which has the following formula:
\begin{equation}
\label{eq:featurematchingloss}
    \mathcal{L}_{FM}(G;D) = \mathbb{E}_{(x,ms)} \left[ \sum_{i=1}^{T} \frac{1}{N_i} \left\lVert D^i(x) - D^i(G(ms)) \right\rVert_1 \right], 
\end{equation}
where $D^i(\cdot)$ and $N_i$ denote the features and the number of features in the $i$-th layer of the discriminator, respectively.

For architectural choices, GAN-based vocoders focus on injecting inductive biases to generate audio waveforms. MelGAN \cite{kumar2019melgan} adds residual blocks with dilations in the generator to model the long-range correlation among the audio time steps and proposes a multi-scale architecture for the discriminator to model the different frequency ranges of the audio. Based on the idea of the multi-scale discriminator, HiFi-GAN \cite{hifigan} proposes a multi-period discriminator to model the diverse periodic patterns within the audio waveforms. To preserve high-frequency content, Fre-GAN \cite{fre-gan} employs the Discrete Wavelet Transform (DWT) to downsample and learn spectral distributions across multiple frequency bands. Unlike traditional approaches like Average Pooling (AP), DWT efficiently decomposes the signal into low-frequency and high-frequency sub-bands. BigVGAN \cite{lee2022bigvgan} introduces a periodic activation function called snake function along with an anti-aliased representation to reduce the high-frequency artifacts in the synthesized audio.


\subsubsection{GAN-based Neural Audio Codec}
Given that many neural audio codecs employ a GAN architecture, they can be effectively discussed within the context of GAN-based vocoders. In contrast to speech tokenizers, the decoder in the codec is leveraged as the vocoder \cite{encodec,soundstream}. Polyak \textit{et al.} \cite{disentangledvocoder} utilizes HiFi-GAN \cite{hifigan} as the vocoder backbone and proposes to disentangle the input features of a vocoder into distinct properties~\cite{disentangledvocoder}, which include semantic tokens, pitch tokens, and speaker embeddings. Such a design choice enables the codec to better perform on pitch and speaker-related tasks such as voice conversion and $F_0$ manipulation.

\subsubsection{Other Types of Vocoder}
The variety of vocoders is not restricted to the ones mentioned earlier, as those are the ones commonly employed in SpeechLMs. This section briefly outlines other potential vocoder types that are seldom explored as a component in SpeechLMs.

\textbf{Pure Signal Processing Vocoder.} Pure signal processing vocoders are traditional methods relying on deterministic algorithms rather than deep learning models to synthesize speech \cite{signalprocessingvocoder1,signalprocessingvocoder2}. However, this kind of vocoder introduces noticeable artifacts in the synthesized audio and is rarely used.

\textbf{Autoregressive Vocoder.} Autoregressive vocoders generate audio waveforms one sample at a time, with each sample conditioned on the previously generated samples \cite{oord2016wavenet}. This approach allows for high-quality audio synthesis due to its sequential nature and the ability to capture intricate temporal dependencies within the audio signal. However, the sequential generation process can be computationally expensive and time-consuming, making autoregressive models less efficient compared to parallelized methods like GAN-based vocoders.


\textbf{Flow-based Vocoder.} Flow-based vocoder aims to establish a series of invertible transformations that map a simple distribution, such as a Gaussian, to the complex distribution of audio samples. This mechanism allows for efficient sampling and density evaluation, enabling the model to synthesize audio in parallel rather than sequentially, which significantly enhances both speed and quality \cite{prenger2019waveglow}. Compared to GAN-based vocoders, Flow-based vocoders typically need more parameters and memory to train the model, which hinders them from being effectively utilized \cite{kumar2019melgan}.


\textbf{VAE-based Vocoders.} Variational Autoencoders (VAEs) are powerful generative models that learn to encode input data into a compressed latent space while allowing for the reconstruction of the original data \cite{vq-vae,vaevocoder1}. However, VAE is seldom explored as the underlying architecture of vocoders.

\textbf{Diffusion-based Vocoder.} Diffusion models have emerged in recent years as a powerful class of generative models that can be used for high-fidelity speech synthesis. They work by gradually adding noise to the input data (e.g. audio waveforms) to create a sequence of increasingly noisy representations, then learning to reverse this process to generate new samples \cite{diffusionvocoder1diffwave,diffusionvocoder2wavegrad,diffusionvocoder3priorgrad}. For instance, DiffWave \cite{diffusionvocoder1diffwave} uses Denoising Diffusion Probabilistic Models (DDPM) to synthesize audio.

In the following paragraph, we present the notation of HiFi-GAN \cite{hifigan} as it is the most used vocoder in SpeechLMs. HiFi-GAN synthesizes high-fidelity audio waveforms from mel-spectrograms or speech tokens using a generator-discriminator framework. The generator \( G(\mathbf{s}; \theta_G) \) maps a sequence of speech tokens \(\mathbf{s}\) to an output audio waveform \(\mathbf{a}\), i.e.,
\begin{equation}
    \mathbf{a} = Vo(\mathbf{s}; \theta_{Vo}) = G(\mathbf{s}; \theta_G),
\end{equation}
where \( Vo \) denotes the vocoder function and \( \theta_{Vo} = \theta_G \) are its parameters. HiFi-GAN employs multi-period and multi-scale discriminators, \( D_{MPD}(\mathbf{a}; \theta_{MPD}) \) and \( D_{MSD}(\mathbf{a}; \theta_{MSD}) \), to distinguish real from generated audio during adversarial training. At inference, only the generator \( G \) is used to efficiently reconstruct speech waveforms.

\section{Training Recipes}
\begin{table} 
    \centering
    \rowcolors{1}{white}{lightblue}
    \caption{A summary of popular datasets used in the pre-training and instruction-tuning phase of SpeechLMs. * means it is the speech version of the text dataset synthesized using TTS. S2ST and S2TT represent speech-to-speech translation and speech-to=text translation, respectively.}
\scalebox{0.74}{
    \begin{tabular}{lllll}
        \toprule
        Dataset  & Type & Phase & Hours & Year\\
        \midrule
        LibriSpeech \cite{librispeechASR}     & ASR    & Pre-Training & 1k & 2015\\
        Multilingual LibriSpeech \cite{MultilingualLibriSpeech}   & ASR   & Pre-Training & 50.5k & 2020\\
        LibriLight \cite{librilightASR}    & ASR    & Pre-Training & 60k & 2019\\
        People dataset \cite{peoplesdatasetASR}    & ASR    & Pre-Training & 30k & 2021\\
        VoxPopuli \cite{voxpopuliASR}    & ASR    & Pre-Training & 1.6k & 2021\\
        Gigaspeech \cite{chen2021gigaspeechASR}    & ASR    & Pre-Training & 40k & 2021\\
        Common Voice \cite{commonvoiceASR}    & ASR    & Pre-Training & 2.5k & 2019\\
        VCTK \cite{vctk}  & ASR & Pre-Training & 0.3k & 2017\\
        WenetSpeech \cite{zhang2022wenetspeechASR}    & ASR    & Pre-Training & 22k & 2022\\
        LibriTTS \cite{libritts}    & TTS    & Pre-Training & 0.6k & 2019\\
        CoVoST2 \cite{wang2020covost2}    & S2TT    & Pre-Training & 2.8k & 2020\\
        CVSS \cite{cvss-ST}    & S2ST    & Pre-Training & 1.9k & 2022\\
        VoxCeleb \cite{voxceleb} & Speaker Identification & Pre-Training & 0.4k & 2017\\
        VoxCeleb2 \cite{voxceleb2} & Speaker Identification & Pre-Training & 2.4k & 2018\\
        Spotify Podcasts \cite{spotifypodcastdataset}    & Podcast    & Pre-Training & 47k & 2020\\
        Fisher \cite{fisherdialoguedataset}    & Telephone conversation    & Pre-Training & 2k & 2004\\
        SpeechInstruct* \cite{speechgpt}    & Instruction-following    & Instruction-Tuning & - & 2023\\
        InstructS2S-200K* \cite{llamaomni}    & Instruction-following    & Instruction-Tuning & - & 2024\\
        VoiceAssistant-400K* \cite{Xie2024MiniOmni} & Instruction-following    & Instruction-Tuning & - & 2024\\
        \bottomrule
    \end{tabular}
}
    \label{tab:datasets}
\end{table}

\label{sec:trainingRecipes}
In this section, we categorize and summarize the commonly used training recipes found in recent SpeechLM papers. This includes an overview of the types of features modeled in SpeechLMs, the various training stages along with the techniques employed in each stage, and the different paradigms for generating speech.

\subsection{Features Modeled}
\label{sec:featuresmodeled}
The features modeled refer to the types of features or tokens outputted by the speech tokenizer and modeled by the language model component within a SpeechLM. These features play a crucial role in determining the capabilities and performance of SpeechLMs. Different features model the speech waveforms from different aspects. In this section, we summarize commonly used features in SpeechLMs and  
focus on how different features affect the performance of SpeechLMs.
Based on recent developments, we can categorize the features modeled by SpeechLMs into two main types: discrete features and continuous features.

\subsubsection{Discrete Features}
\label{sec:discretefeatures}
Discrete features (or discrete tokens) refer to quantized representations of speech signals that can be represented as distinct, countable units or tokens. These features are typically derived from speech signals through various encoding and quantization processes, resulting in a finite set of possible values. Discrete features are the most used features by SpeechLMs as they can be represented as tokens and be modeled exactly the same as the text tokens within a TextLM.

Most SpeechLMs only employ \textit{semantic tokens} (generated by semantic understanding tokenizers, Section \ref{sec:semanticUnderstandingTokenizer}) to represent speech, as semantic information plays the most crucial role in spoken communication. GSLM \cite{OnGenerative}, the first-ever SpeechLM, compares three tokenizers, which include Contrastive Predictive Coding (CPC) \cite{cpc}, wav2vec 2.0 \cite{wav2vec2.0}, and HuBERT \cite{hsu2021hubert}. It concludes that HuBERT performs the best on various tasks such as speech resynthesis and speech generation. A large number of works follow this setting and use HuBERT as the speech tokenizer \cite{textuallypretrainSLM,spiritlm,speechgpt}. AudioPaLM \cite{audiopalm} experiments the choice between w2v-bert \cite{w2v-bert}, USM-v1 \cite{zhang2023googleusm}, and USM-v2 \cite{audiopalm} (a modified version of USM-v1), and it concludes that USM-v2 is the best-performing speech tokenizer on ASR and Speech Translation (ST) tasks.

Although semantic tokens excel at generating semantically meaningful speech because of the modeling of the contextual information within speech waveforms, researchers find out that the speech generated solely upon semantic tokens lacks expressive information such as prosody and different pitches or timbres \cite{expressodataset,spiritlm}. To conquer this limitation, \textit{paralinguistic tokens} can be integrated into the modeling process to capture expressive information with speeches. pGSLM \cite{prosodyawareSLM} proposes to use the fundamental frequency (F0) and unit duration as prosody features to complement the HuBERT semantic tokens, and trains a multi-stream transformer language model to predict the semantic tokens, pitch (F0), and unit duration separately. 
Similarly, SPIRIT-LM \cite{spiritlm} complements the HuBERT semantic tokens with pitch and style tokens \cite{sonarspeechprop}. This incorporation of extra acoustic tokens allows SpeechLMs to more effectively capture expressive elements without significantly compromising semantic understanding \cite{spiritlm}.

Another type is \textit{acoustic tokens}, which are tokens aiming to capture essential acoustic features to reconstruct high-fidelity speech, primarily obtained from neural audio codec models (Section \ref{sec:acousticGenerationTokenizer}). Some studies directly model the codec tokens in a language model, which is often regarded as Codec Language Models (CodecLMs). For example, Viola \cite{violaCodecLM} trains a CodecLM capable of performing ASR, TTS, and Machine Translation. NTPP \cite{ntpp} trains on VQ-VAE \cite{vq-vae} tokens for modeling dual-channel spoken dialogue data.

\textbf{Discussion.} Different types of tokens influence the speech quality of SpeechLMs in different ways, often resulting in trade-offs \cite{audiolm}. For example, while semantic tokens align well with text and excel in producing semantically coherent speech, the generated speech often lacks acoustic details, such as high-frequency information. Recovering and enhancing these details typically requires post-processing, like a diffusion model, which significantly increases the model's latency. Conversely, acoustic tokens can facilitate the generation of high-fidelity audio but often struggle with inaccuracies in content generation \cite{zhang2023speechtokenizer}. Researchers have tried two ways to balance these trade-offs. The first involves combining semantic and acoustic tokens into a single sequence. AudioLM \cite{audiolm} proposes a hierarchical modeling scheme that first models semantic tokens from w2v-bert \cite{w2v-bert} and then uses these tokens to predict acoustic tokens from SoundStream \cite{soundstream}, which ultimately generates speech. However, this kind of approach increases sequence length, which increases modeling complexity. The second strategy leverages \textit{mixed tokens} (Section \ref{sec:mixedTokenizer}) to jointly model semantic and acoustic information, showing promising results in Moshi \cite{defossezmoshi} and SpeechGPT-Gen \cite{speechgpt-gen}.



\begin{table*}
    \centering
    \caption{Four different methods of modeling speech and text tokens.}
    \rowcolors{1}{white}{lightblue}
    \begin{tabular}{p{3.5cm} p{6.5cm} p{6cm}}
        \toprule
        Modeling Method  & Example & Explanation\\
        \midrule
        Speech-only     & [SPEECH] S12 S34 S33 ... S11 S59    & Only the speech sequence is provided.\\
        Text-only       & [TEXT] A quick brown fox jumps over a lazy dog.    & Only the text sequence is provided.\\
        Concatenated speech-text     & [SPEECH] S12 S34 S33 ... S11 S59 [TEXT] A quick brown fox jumps over a lazy dog.    & The speech sequence and text sequence are concatenated together.\\
        Alternating speech-text       & [SPEECH] S12 S34 S33 [TEXT] brown fox jumps over a lazy [SPEECH] S11 S59    & The sequence is interleaved with speech and text tokens.\\
        \bottomrule
    \end{tabular}
    \label{tab:modelingtypes}
\end{table*}

\subsubsection{Continuous Features}
\label{sec:continuousfeatures}
Continuous features (or continuous tokens), in contrast to discrete features, are unquantized, real-valued representations of speech signals that exist on a continuous scale (continuous tokens). Continuous features can include spectral representations like mel-spectrograms or latent representations extracted from neural networks. Spectron \cite{spectrogramSLM} performs speech continuation by predicting the spectrograms frame-by-frame. 
Mini-Omni \cite{Xie2024MiniOmni} and SLAM-Omni \cite{SLAM-Omni} extract intermediate representations from a frozen Whisper encoder as input for the SpeechLM, whereas LauraGPT \cite{Du2023LauraGPT} employs an audio encoder trained alongside the language model to derive latent representations from input speech. Continuous features can capture fine-grained, nuanced aspects of speech that may be lost in discretization processes. However, utilizing these features often necessitates modifying the off-the-shelf training pipeline of language models, as traditional text-based models are built to handle discrete units. Moreover, continuous features demand more storage capacity compared to their discrete counterparts.

\subsection{Training Stages}
\label{sec:trainingstages}
Training a SpeechLM involves training the three main components: speech tokenizer, language model, and vocoder. Similar to TextLMs, the key to training SpeechLMs lies in effectively modeling speech continuation, which is primarily the responsibility of the language model. The speech tokenizer and vocoder usually rely on established methods and are trained using distinct training datasets specific to each SpeechLM approach. Therefore, This section reviews the main techniques used to train the language model component. Following TextLMs, we divide the training process of SpeechLMs into three stages, including pre-training, instruction-tuning, and post-alignment.

\subsubsection{Language Model Pre-Training}
The pre-training of the language model in SpeechLMs is a critical phase that significantly influences the model's ability to generate coherent and contextually relevant speech. This phase typically involves training the language model to autoregressively predict the next token on a large corpus of speech tokens. The primary objective during this stage is to learn the statistical patterns and dependencies inherent in the speech data, enabling the model to predict the next token in a sequence based on the preceding context.

\textbf{Training data.} SpeechLMs pre-training mainly leverages large-scale open-sourced speech data. Commonly used datasets include those for ASR \cite{librispeechASR,librilightASR,peoplesdatasetASR,voxpopuliASR}, TTS \cite{libritts}, ST \cite{cvss-ST,voxpopuliASR}, podcasts \cite{spotifypodcastdataset}, and dialogues \cite{fisherdialoguedataset}. Table \ref{tab:datasets} includes popular datasets used in the pre-training stage. Some datasets consist solely of speech data, while others include both speech and corresponding text transcripts. The inclusion of text transcripts can enhance the model's representation by allowing it to learn the relationship between spoken language and its written form, which will be discussed later.

\textbf{Cold Initialization.} Some SpeechLMs use cold initialization during the pre-training phase, where model parameters are initialized randomly. The pioneering SpeechLM---GSLM \cite{OnGenerative}---trained a transformer \cite{transformer} from scratch to serve as the language model. This study demonstrated the effectiveness of the SpeechLM pipeline and compared performance across various speech tokenizer options. They found that HuBERT \cite{hsu2021hubert} outperformed CPC \cite{cpc} and wav2vec 2.0 \cite{wav2vec2.0} in understanding speech content and generating natural speech. SUTLM \cite{jointmodelingSpeechText} also uses a transformer as the language model. They studied the critical problem of jointly modeling speech and text tokens by comparing four different modeling methods: speech-only, text-only, concatenated speech-text, and alternating (interleaving) speech-text. They showed that the setting of alternating speech-text performs the best in cross-modal evaluations. Table \ref{tab:modelingtypes} illustrates the four modeling methods.

Some works leverage a different architecture from the standard transformer. These models are typically trained from scratch when the architecture deviates from a standard transformer or TextLM too much. For example, pGSLM \cite{prosodyawareSLM} proposes a multi-stream transformer language model (MS-TLM) that takes multiple streams of input and predicts multiple streams of output to generate speech units, duration, and pitch embeddings simultaneously. dGSLM \cite{generativedialog2channel} introduced a dialogue transformer language model (DLM) to jointly model the dialogue speech data from the two speakers. To enable the listening ability of SpeechLMs while speaking, LSLM \cite{LMListenWhileSpeaking} proposes to attach a streaming self-supervised learning (SSL) Encoder to an autoregressive token-based TTS Model.

\textbf{Continued Pre-Training.} In contrast to cold initialization, continued Pre-Training involves initializing the language model with pre-trained weights from a TextLM and then adapting it to handle speech tokens. This approach leverages the linguistic knowledge embedded in TextLMs, allowing for more efficient and effective training of SpeechLMs. Hassid \textit{et al.} \cite{textuallypretrainSLM} found that starting with a textually pre-trained language model (OPT \cite{opt} and LLaMA \cite{llama}) can enhance the model's convergence rate and significantly improve its speech understanding capabilities. They also demonstrated that while training from text-pretrained checkpoints outperforms cold initialization, training from image-pretrained checkpoints yields poorer results compared to cold initialization. This indicates that not all pre-trained checkpoints are equally effective. Additionally, AudioPaLM \cite{audiopalm} trained the SpeechLM using PaLM and PaLM-2 \cite{Chowdhery2023Palm,palm2}, showing that the SpeechLM benefits from both an increased size of the pre-trained checkpoint and a larger training dataset.

The performance of SpeechLMs can be further enhanced by \textbf{aligning the text and speech modality representations.} Some works align the text and speech representations in a \textbf{single sequence}. SPIRIT-LM \cite{spiritlm} found that continually pretraining on TextLM checkpoints using interleaving text and speech tokens can significantly boost the model's performance on speech understanding and generation. Additionally, their visualizations demonstrate that the similarity between text and speech features is notably higher in models trained with interleaved token sequences compared to those trained without this approach. 
Spectron \cite{spectrogramSLM} solve the text-speech representation alignment problem by jointly supervising multiple objectives. Specifically, the input speech prompt is first transcribed into its text tokens, and then the model predicts the text token response. Finally, the text response is synthesized to output speech. SpeechGPT \cite{speechgpt} also adopts this concept but applies it during the instruction-tuning phase.
Some other methods perform \textbf{multi-sequence} representation alignment. This approach simultaneously generates the text sequence and speech sequence(s). For example, Llama-Omni uses the LLM output hidden state to decode text tokens and generate discrete speech tokens simultaneously. Mini-Omni \cite{Xie2024MiniOmni} generates a single sequence of text tokens and seven sequences of acoustic tokens in parallel, all aligned at the sentence level. Similarly, Moshi \cite{defossezmoshi} generates one sequence of text tokens, one sequence of semantic tokens, and seven sequences of acoustic tokens in parallel, which are aligned at the word level.

\textbf{Discussion.} The primary goal of aligning text and speech representations is to leverage the strengths of text-based models to enhance speech-based models. Researchers have found that training a SpeechLM is significantly more challenging than training a TextLM. This difficulty arises because text serves as a concentrated form of knowledge, while speech requires models to independently learn the rules of spoken language. Aligning text and speech representations has demonstrated effectiveness, but it involves various trade-offs. First, text primarily conveys semantic information, which can improve a SpeechLM's semantic modeling capabilities but may compromise its ability to capture paralinguistic features, such as tone and emotion, during alignment. Second, there are two main inference approaches for the aligned models: text-present and text-independent. Text-present inference decodes text and speech simultaneously, which may increase latency but enhances the SpeechLM's reasoning abilities \cite{Xie2024MiniOmni} and reduces possible hallucinations \cite{defossezmoshi}. Conversely, text-independent inference is more efficient but may lack stability. Furthermore, the question of whether to incorporate text modality to enhance SpeechLM performance remains an open question, especially considering that humans typically acquire spoken language skills before mastering written language.

\subsubsection{Language Model Instruction-Tuning} Instruction-tuning refers to the process of fine-tuning SpeechLMs to follow specific instructions to perform a wide range of tasks. This phase is crucial for enhancing the pre-trained model's generalization capabilities and making it more adaptable to diverse applications. Therefore, the key focus is on creating effective instruction-following datasets.

Several approaches have been proposed to construct instruction-following datasets for SpeechLMs. SpeechGPT \cite{speechgpt} and SpeechGPT-Gen \cite{speechgpt-gen} propose a two-stage instruction-tuning, including cross-modal instruction fine-tuning and chain-of-modality instruction fine-tuning. In the first stage, instruction data are generated based on ASR datasets by appending the instruction to paired ASR data, asking the model to convert speech into text. Similarly, paired data is also used to create instruction data for performing TTS. In the second stage, they construct a speech-in-speech-out dataset by transforming a text-based instruction-following dataset using TTS. Llama-Omni \cite{llamaomni} also creates instruction-following data by synthesizing text-based datasets, adhering to specific constraints. First, they transform the input text prompt into a format that mimics natural speech patterns. Next, they discard the original text response and employ a TextLM to generate answers to the converted prompts, ensuring these responses also follow natural speech patterns. Finally, they synthesize the prompt/response pairs using TTS. COSMIC \cite{pan2023cosmicSpeechInstructionTuning} constructed speech QA data by asking GPT-3.5 to generate question-answer pairs based on the transcriptions of English TED talk speeches. They showed the model trained on their proposed speech QA dataset can generalize to unseen tasks such as speech-to-text translation using in-context learning.

\subsubsection{Language Model Post-Alignment}
Post-alignment is the critical process of refining a language model's behavior to align with human preferences, ensuring that its outputs are both safe and reliable. This stage is typically regarded as the final phase of language model training. It often employs techniques like Reinforcement Learning from Human Feedback (RLHF), specifically methods such as Proximal Policy Optimization (PPO) \cite{ppo} and Direct Preference Optimization (DPO) \cite{dpo}. 

Post-alignment in SpeechLMs focuses on addressing the unique challenges inherent in the speech interaction pipeline. Align-SLM \cite{alignslm_RLHF_RLAIF} identifies that SpeechLMs often generate inconsistent semantic content when given the same prompt. It tackles this by using a TextLM to choose the preferred response from SpeechLMs after transcribing them via ASR and then aligning those preferences using DPO. On the other hand, SpeechAlign \cite{Zhang2024SpeechAlign} concentrates on the acoustic quality of SpeechLMs. It observes that differences between the ``golden" speech tokens and those generated by the language model lead to subpar acoustic quality in the generated speech as the vocoder synthesizes speech from generated tokens during inference. To mitigate this, it employs various optimization techniques to align the language model's output with the distribution of the ``golden" tokens. Despite its importance, the post-alignment of SpeechLMs remains under-explored. A critical application of post-alignment is to mitigate the safety risks associated with generative models. Thus, future research should prioritize identifying and addressing the unique safety challenges posed by SpeechLMs (see Section \ref{chall:safetyrisks}).

\subsection{Speech Interaction Paradigm}
\label{sec:trainingparadigm}
Most approaches covered in earlier sections follow the traditional generation paradigm of SpeechLMs, which involves taking a predefined input sequence and generating a complete response. However, this approach does not reflect the natural flow of voice interactions. For instance, during a conversation, one person may interrupt another, switching from listening to speaking. Additionally, a person might choose not to respond if the other is engaged in a conversation with someone else. Based on these observations, we identify two key aspects of advanced speech interaction skills for SpeechLMs: real-time interaction and interactive period recognition.

\textbf{Real-time Interaction} of SpeechLMs involves the advanced handling of conversation data from two or more people, and it can be understood through several progressive stages. The initial stage is the adoption of \textbf{streaming tokenizers and vocoders}, which eliminate the need for the language model to wait for complete speech encoding before processing. This architecture enables immediate, low-latency responses to user queries, marking a significant improvement over the traditional interaction paradigm. Nonetheless, while this streaming approach supports basic real-time interaction, it remains insufficient for capturing the more sophisticated interaction patterns observed in natural conversation. The next frontier is \textbf{full-duplex modeling}, which allows SpeechLMs to support simultaneous bidirectional communication—specifically, the ability to handle interruptions initiated by either the user or the model. It mainly includes two features: 1) User interruption, where models can be interrupted and respond appropriately to new instructions during a conversation, and 2) Simultaneous response, enabling models to process input and generate output concurrently. Achieving this requires the joint modeling of both user and model audio streams. dGSLM \cite{generativedialog2channel} employs a separate transformer for each participant in two-speaker dialogues, with cross-attention layers capturing speaker interactions. Most methods, however, rely on a single language model. NTPP \cite{ntpp} employs a ``next-token-pair prediction" approach with a decoder-only Transformer to predict tokens for both channels. Moshi \cite{defossezmoshi} concatenates user input and model response channels data, using an RQ-Transformer to process the data together. LSLM \cite{LMListenWhileSpeaking} focuses on modeling one speaker's speech using a decoder-only Transformer, integrating a streaming SSL encoder to fuse listening and speaking channel embeddings.

\textbf{Interactive Period Recognition (IPR)} refers to the ability to recognize whether the users are interacting with it or not. SpeechLMs should provide response during the interactive period and remain silent during the non-interactive period. IPR is essential for creating a natural conversational flow, allowing the model to avoid unnecessary interruptions. It is crucial for situations where a small group of users is having a discussion, as the SpeechLM needs to discern when to join in and when to stay silent. Additionally, it is important for the model to learn when to disregard instructions when users are not speaking at it. One approach to achieving IPR is through a Voice Activity Detection (VAD) module. MiniCPM-o 2.6 \cite{MiniCPM-0-2.6} integrates a VAD module to ensure the model responds only when the input audio surpasses a predefined VAD threshold. Inputs below this threshold are considered noise and ignored. VITA \cite{VITAtencentOmniMLLM} takes a different approach by training the SpeechLM to distinguish between query speech and non-query audio. The model learns to output an end-of-sequence token to terminate its response when non-query audio is detected.


\section{Downstream Applications}
\label{sec:downstreamApps}


\begin{table*}[t!]
    \centering
    \caption{Examples of the various capabilities of SpeechLMs.}
    \begin{tabular}{p{3cm} p{6.5cm} p{6.5cm}}
        \toprule
        Task  & Input & Output\\
        \midrule
        \rowcolor{lightblue} \multicolumn{3}{c}{\textbf{Semantic-related Applications}} \\
        Spoken Dialogue     & Speech(I am so angry right now!)    & Speech(Sorry to hear that. What happened?)\\
        Speech Translation  & Speech(Good morning my friend!) Please translate the speech into French.    & Speech(Bonjour mon ami !)\\
        Automatic Speech Recognition  & Speech(A river is on your left side.) Please perform automatic speech recognition to the speech   & The transcribed speech is: A river is on your left side.\\
        Keyword Spotting  & Speech(How's your day? (Speaking to somebody else) Hey, Siri.) Please detect if the given speech contains any pre-defined keywords.   & Keyword detected: ``Hey, Siri."\\
        Text-to-Speech Synthesis  & Please synthesize the following text into speech: ``A river is on your left side."    & Sure. Here is the synthesized speech: Speech(A river is on your left side.)\\
        Intent Classification  & Speech(Can you help me book a flight?) Please detect the intent of the given speech.    & Intent detected: ``Book a flight".\\
        Slot Filling  & Speech(I want to fly from New York to San Francisco on June 5th.) Please fill the following slots based on the given speech: departure city, destination city, and date    & Sure. Here are the filled slots. Departure city: New York. Destination city: San Francisco. Date: June 5th.\\
        Query by Example Spoken Term Detection  & Speech(Can you retrieve the audio documents that contain the word ``happy"?) [A batch of audio documents]    & Sure. I have found documents 2 and 4 contain the word happy.\\
        \midrule
        \rowcolor{lightblue} \multicolumn{3}{c}{\textbf{Speaker-related Applications}} \\
        Speaker Identification  & The following speech comes from Lucy: ``[Lucy's speech]". The following speech comes from Bob: ``Bob's speech". Question: Who is the speaker of the following speech [Unknown speech]    & The speaker of the following speech: ``[Unknown speech]” is Bob.\\
        Speaker Verification  & Here are two speeches. Speech 1: [Speech 1]. Speech 2: [Speech 2]. Do they come from the same speaker?    & No, they do not come from the same speaker.\\
        Speaker Diarization  & [An audio file of the meeting] Please provide the speaker diarization output in the following format: [Start Time] - [End Time]: Speaker [Speaker].    & Sure. Here is the speaker diarization result: 00:00 - 00:15: Alice, 00:16 - 00:40: Bob, 00:41 - 01:00: Carol, 01:01 - 01:20: Alice, 01:21 - 01:45: Bob and Carol (overlapping speech), 01:46 - 02:00: Carol.\\
        Voice-Conditioned Speech Generation  & Can you use a robotic voice to say "Hello world!" to me?   & Sure. Speech([In robotic voice] Hello world!)\\
        \midrule
        \rowcolor{lightblue} \multicolumn{3}{c}{\textbf{Paralinguistic Applications}} \\
        Emotion Recognition  & Speech(What are you doing) What emotion does the speech convey?   & It seems that the speaker is pretty shocked and angry about something.\\
        Speech Separation  & Speech([Two overlapping speeches]) Can you separate the two speeches?  & Sure. Here is the speech separation result. Speech 1: Speech(Should I go for this one?). Speech 2: Speech(That one seems good.)\\
        Paralinguistics-Enhanced Generation  & Speech(Please generate a speech that sounds happy.)    & Speech([In happy mood] My friend just gave me a candy!!)\\
        \bottomrule
    \end{tabular}
    \label{tab:downstream_apps}
\end{table*}

Unlike traditional speech systems like ASR and TTS, which usually focus on specific tasks, SpeechLMs function as generative foundation models. They can handle a diverse array of speech-only, text-only, and multi-modal tasks by following various instructions. In this section, we explore the primary downstream applications of SpeechLMs. The tasks discussed here primarily consist of traditional speech-related tasks, along with some that are unique to SpeechLMs. In contrast to TextLMs, which generate text containing only semantic information, SpeechLMs can model both semantic and paralinguistic information---such as pitch and timbre---making them more powerful models. We thereby categorize the downstream applications of SpeechLMs into three main classes: semantic-related applications, speaker-related applications, and paralinguistic applications. Table \ref{tab:downstream_apps} provides an example for each downstream task.

\subsection{Semantic-Related Applications}
Semantic-related applications encompass key tasks that facilitate meaningful interactions between humans and machines. These applications require SpeechLMs to understand the semantic meaning of the input and generate responses that are not only contextually relevant but also logically coherent. The primary Semantic-related applications of SpeechLMs are as follows.


\textbf{Spoken Dialogue.} Spoken dialogue is the most natural application of SpeechLMs. Spoken dialogue systems are designed to facilitate natural conversations between humans and machines in spoken format. They can engage users in interactive exchanges, understanding and generating responses based on the context of the conversation. Unlike TextLMs, SpeechLMs are able to perform conversations with humans directly in speech, which is a more natural way of communication. Note that SpeechLMs can not only perform speech-only dialogues but also perform cross-modal dialogues, such as taking texts as input and responding in speech format.

\textbf{Speech Translation.} Speech translation (ST) is the process of converting spoken language from one language to another. Similar to Spoken dialogue, SpeechLMs can perform ST in both single-modal and cross-modal settings. Specifically, the input and output of the ST task can be either in text or speech format.

\textbf{Automated Speech Recognition.} Automatic speech recognition (ASR) enables systems to convert spoken language into text. The input of ASR is a speech waveform, and the system outputs the transcription in textual form. For SpeechLMs, the input would be a combination of the speech waveform and the instruction to tell the model to perform ASR on the given speech.

\textbf{Keyword Spotting.} Keyword spotting can be considered a special type of ASR, where its primary objective is to identify specific words or phrases within continuous speech. While traditional ASR systems aim to transcribe entire spoken utterances into text, keyword spotting focuses specifically on identifying and extracting predefined keywords or phrases within continuous speech. The primary application of keyword spotting is to build voice-activated assistants in smart home devices. Those devices are activated when the specific keywords are triggered. Therefore, although SpeechLMs are capable of spotting and understanding more than just a couple of words, keyword spotting can be used to efficiently trigger SpeechLMs to respond to user inputs.

\textbf{Text-to-Speech Synthesis.} Text-to-speech synthesis (TTS) enables systems to synthesize written text into spoken language. In contrast to ASR, TTS takes text as input and outputs the converted speech waveform. Similarly, the input of the SpeechLMs would be a combination of the text to synthesize and the instruction, and the output is the synthesized speech.

\textbf{Intent Classification.} Intent classification is a critical task that identifies the underlying intention behind a user's input speech. The AI system can then perform certain actions based on the identified user intent (e.g., book a flight). Intent classification is particularly important in applications such as virtual assistants, customer service bots, and interactive voice response systems. To perform Intent Classification, it is more natural for SpeechLMs to take speech inputs and classify the results in text since it is easier to parse and classify the intent classification result in text than speech.

\textbf{Slot Filling.} Slot filling is an important task in spoken language understanding that involves identifying and extracting specific pieces of information from user inputs into predefined classes, such as intents, entities, and parameters that are essential for completing a task. For example, slot filling extracts the phrase ``I want to fly from New York to San Francisco on June 5th." into distinct slots like ``departure city" (New York), ``destination city" (San Francisco), and ``date" (June 5th). Similar to Intent Classification, it is more natural for SpeechLMs to take speech inputs and extract the pieces in texts.

\textbf{Query by Example Spoken Term Detection.} Another spoken term detection task is query by example spoken term detection (QbE-STD), which allows users to identify specific spoken terms or phrases within a larger audio stream by providing an example of the desired term. Unlike traditional keyword spotting methods that rely on predefined lists of keywords, QbE-STD leverages the flexibility of example-based querying, enabling users to specify their search terms through audio samples.

\subsection{Speaker-Related Applications}
Speaker-related applications refer to the tasks that involve the processing of information related to speaker identity. It could involve classification tasks such as identifying, verifying, and distinguishing individual speakers based on their unique vocal characteristics, as well as generation tasks such as maintaining or modifying the timbre of a given speech. While we acknowledge that voice characteristics can be considered paralinguistic information, we believe that speaker-related applications are unique because they enable SpeechLMs to function in complex scenarios such as participating in multi-speaker conversations. In this section, we survey common speaker-related applications of SpeechLMs. 

\textbf{Speaker Identification.} Speaker identification is the process of recognizing a person's identity based on their voice characteristics. It is a multi-class classification of a given speech as input. SpeechLMs can perform this task by taking an input speech and outputting the classification result in text or speech format. Moreover, SpeechLMs can also identify different speakers implicitly. Specifically, it can chat with multiple speakers at the same time, distinguishing the words from different speakers and responding to each speaker appropriately.

\textbf{Speaker Verification.} Speaker verification involves determining whether the speakers of a pair of speeches match with each other. Unlike speaker identification, which is a multi-class classification process, speaker verification is a binary classification process.

\textbf{Speaker Diarization.} Speaker diarization is the process of partitioning an audio stream into segments according to the identity of the speakers. It predicts ``who is speaking when" for each timestamp \cite{yang2021superb}. A natural way to integrate speaker diarization into SpeechLMs is to have the model generate the transcript of each audio segment along with the identification of the speaker.

\textbf{Voice-Conditioned Speech Generation.} Voice-conditioned speech generation involves synthesizing speech based on the vocal characteristics of a specific speaker. This could involve voice cloning and voice conversion. Voice cloning utilizes a sample of the speaker's voice as a reference, enabling the model to reproduce the speaker's timbre when generating speech from input text. Voice conversion, on the other hand, modifies an existing speech signal to sound like it was produced by a different speaker while retaining the original content. Additionally, instead of giving the target vocal characteristics, SpeechLMs should also be able to adapt their output timbre based on various speech or text instructions.

\subsection{Paralinguistic Applications}
Paralinguistics refers to the non-verbal elements of communication that accompany spoken language. It encompasses various vocal attributes that convey meaning beyond the actual words spoken. These elements can significantly influence how messages are interpreted and understood. The key elements of paralinguistics include pitch, timbre, column, rate of speech, pauses, etc. Since combining elements in paralinguistics in different ways can result in a speech with different emotions, we include emotion-related tasks as paralinguistic applications as well.

\textbf{Emotion Recognition.} Emotion recognition task involves identifying and classifying the emotion carried by a given speech into predefined classes. Similar to speaker identification, SpeechLMs are capable of not only directly performing this task but also implicitly recognizing users' emotions through their speech queries and responding accordingly.

\textbf{Speech Separation.} Speech separation refers to the process of isolating individual speech signals from a mixture of sounds, such as when multiple speakers are talking simultaneously. When separating the input speech, SpeechLMs can not only output the contents of each person in speech but also in text format (i.e., transcriptions).

\textbf{Paralinguistics-Enhanced Generation.} Paralinguistics-enhanced generation refers to the process of instructing SpeechLMs to produce speech that exhibits specific paralinguistic characteristics. Users can define these characteristics in their prompts, allowing the model to generate speech that aligns with their specifications. Examples of paralinguistics-enhanced generation include synthesizing speech with a specific style, speaking at a fast pace, and even singing. This capability distinguishes SpeechLMs from TextLMs and facilitates a more engaging and interactive form of communication with the AI models.

\section{Evaluations}
\label{sec:evaluation}

\begin{table} 
    \centering
    \rowcolors{1}{white}{lightblue}
    \caption{A summary of popular benchmarks for the evaluation of SpeechLMs. I/O, $A$, and $T$ represent input/output modality, audio, and text, respectively.}
\scalebox{0.75}{
    \begin{tabular}{lllll}
        \toprule
        Name & Eval Type & \# Tasks & Audio Type & I/O\\
        \midrule
        ABX \cite{speechbenchmark2015,speechbenchmark2019,speechbenchmark2021}     & Representation    & 1 & Speech & $A \rightarrow -$\\
        sWUGGY \cite{speechbenchmark2021}    & Linguistic    & 1 & Speech & $A \rightarrow -$\\
        sBLIMP \cite{speechbenchmark2021}    & Linguistic    & 1 & Speech & $A \rightarrow -$\\
        sStoryCloze \cite{textuallypretrainSLM}     & Linguistic    & 1 & Speech & $A/T \rightarrow -$\\
        STSP \cite{spiritlm}     & Paralinguistic    & 1 & Speech & $A/T \rightarrow A/T$\\
        MMAU \cite{mmau_benchmark}     & Downstream    & 27 & Speech, Sound, Music & $A \rightarrow T$\\
        Audiobench \cite{audiobench_benchmark}     & Downstream    & 8 & Speech, Sound & $A \rightarrow T$\\
        AIR-Bench \cite{airbench_benchmark}     & Downstream    & 20 & Speech, Sound, Music & $A \rightarrow T$\\
        SD-Eval \cite{sd-eval_benchmark}     & Downstream    & 4 & Speech & $A \rightarrow T$\\
        SUPERB \cite{dynamicsuperb_benchmark}     & Downstream    & 10  & Speech & $A \rightarrow T$\\
        VoxDialogue \cite{voxdialogue}     & Downstream    & 12 & Speech, Sound, Music & $A \rightarrow T$\\
        Dynamic-SUPERB \cite{dynamicsuperb_benchmark}     & Downstream    & 180  & Speech, Sound, Music & $A \rightarrow T$\\
        SALMON \cite{salmon_benchmark}     & Downstream    & 8 & Speech & $A \rightarrow -$\\
        VoiceBench \cite{VoiceBench}     & Downstream    & 8 & Speech & $A \rightarrow T$\\
        VoxEval \cite{cui2025voxeval}     & Downstream    & 56 & Speech & $A \rightarrow A$\\
        \bottomrule
    \end{tabular}
}
    \label{tab:benchmarks}
\end{table}

Similar to TextLMs, SpeechLMs have a wide range of capabilities, making it challenging to compare different SpeechLMs. Consequently, it's essential to evaluate SpeechLMs from various perspectives to determine their effectiveness. In this section, we review the commonly used methods and benchmarks (Table \ref{tab:benchmarks}) for evaluating SpeechLMs. We categorize these evaluation methods into automatic and human assessments, each containing distinct evaluation aspects.

\subsection{Automatic (Objective) Evaluation}
Automatic evaluation methods are essential for providing quick and consistent assessments of SpeechLMs. These methods typically rely on quantitative metrics that can be computed without human intervention. Below, we outline some of the most commonly used automatic evaluation techniques.

\textbf{Representation Evaluation.} Representation (embedding) is a crucial component in SpeechLMs (and TextLMs). It refers to how input data, such as speech or text, is transformed into a format that the model can understand and process. Effective representation lays a solid foundation for models to understand lexical, syntax, and contextual information, which are vital for generating coherent and contextually relevant outputs.

In the context of SpeechLMs, representation evaluation focuses on how well the model encodes speech features into meaningful vectors. GSLM \cite{OnGenerative} uses \textit{between-speaker ABX score} to measure the embedding similarity. It quantifies how well-separated the phonetic categories are. Specifically, It works by comparing three sound samples: two from the same category (A) and one from a different category (B). The test measures how often the system correctly identifies that two sounds from category A are more similar to each other than one sound from A is to a sound from B. Another way of evaluating representations is through speech resynthesis \cite{OnGenerative}. Specifically, an input speech is encoded into tokens and then synthesized back to speech. Then, word error rate (WER) or character error rate (CER) can be computed on the ASR results of the input and resynthesized speech. This measures the information loss caused by discretizing the input speech into speech tokens, thereby evaluating the robustness of the latent representations.

\textbf{Linguistic Evaluation.} Linguistics, including lexical, syntactic, and semantic evaluation methods, assess the model’s ability to generate and understand the rules for constructing words, sentences, and meaningful contents. These evaluations focus on the correctness and appropriateness of word choices, the grammatical structure of the outputs, and the coherence and relevance of the generated content. In terms of benchmark datasets, sWUGGY \cite{speechbenchmark2021} assesses at the lexical level by determining if the model can distinguish a real word from a (real, non-real) word pair. sBLIMP \cite{speechbenchmark2021} evaluates at the syntactic level by determining if the model can identify the grammatically correct sentence from a (grammatical, ungrammatical) sentence pair. Spoken StoryCloze \cite{textuallypretrainSLM} evaluates semantic comprehension by assessing the model's capability to select the genuine ending of a story from a pair of ending choices. All the evaluation is conducted by comparing the model's negative log-likelihood of the data pair.

\textbf{Paralinguistic Evaluation.} In contrast to linguistic evaluation, paralinguistic evaluation focuses on the non-verbal aspects of communication that accompany speech. Some works choose to utilize paralinguistic tokens alongside semantic tokens to enhance the paralinguistic abilities of SpeechLMs \cite{prosodyawareSLM,spiritlm}, so one way is to evaluate the paralinguistic tokens. pGSLM \cite{prosodyawareSLM} measures the correctness, consistency, and expressiveness of the prosodic tokens. Correctness evaluates the model's ability to generate accurate prosodic profiles by calculating the minimal mean absolute error (min-MAE) of the prosodic tokens from 20 generated samples against the prosodic tokens from the reference, consistency is assessed through the Pearson correlation between the mean values of the prompt prosodic and its generated continuation prosodic tokens, and expressiveness is measured by the standard deviation of the generated prosody token values, with the expectation that it matches the variability of the ground truth. We note that the same metrics can also be applied to other paralinguistic tokens. Instead of evaluating from the token level, SPIRIT-LM \cite{spiritlm} proposes to measure on the perceptual level. They introduced a speech-text sentiment preservation benchmark (STSP), which requires the model to generate a text or speech sequence of tokens that preserves the sentiment of the prompt. A sentiment classifier is used to assess the sentiment in the generated speech. It should be noted that although they only apply the preservation approach to sentiment, this idea can be generalized to other paralinguistic features, such as timbre or prosody.

\textbf{Generation Quality and Diversity.} Quality and diversity are two crucial aspects of model generation. Typically, there is a trade-off between these dimensions when sampling model responses at different temperatures, so GSLM \cite{OnGenerative} suggests using the Area Under the Curve (AUC) with various temperature values. Specifically, AUC on perplexity and VERT are employed to assess these factors, where VERT represents the geometric mean of the ratio of k-grams in the generated speech that repeat at least once. Additionally, the ChatGPT score can be utilized to evaluate the quality of the generated speech. In this process, the generated speech is transcribed using state-of-the-art ASR models and then sent to ChatGPT for quality (and diversity) assessment.

\textbf{Real-Time Interaction Evaluation.} Real-time interaction evaluation involves assessing the ability of SpeechLMs to interact in real-time, which is crucial for those models that facilitate streaming or full-duplex interaction. Current research focuses on evaluating the naturalness and usefulness of speech generated in real-time. dGSLM \cite{generativedialog2channel} examines the naturalness of dialogues between two speakers by introducing different turn-taking events, such as speech segments (Inter-Pausal Unit, IPU), pauses within speech (pause), pauses between speeches (gaps), and overlapping speech (overlap). The generated speech is considered more natural if the statistics of these turn-taking events closely resemble those found in human dialogues. Another method involves speech continuation, where the generated speech is considered more natural if the turn-taking statistics of the speech prompts align closely with those of the subsequent continuations. Recently, Full Duplex Bench \cite{fullduplexbench} serves as a benchmark to evaluate various turn-taking abilities in full duplex SpeechLMs, and Talking Turns \cite{talkingturns} evaluate turn-taking events by training a neural network model to predict the events outputted by the full duplex SpeechLMs. Additionally, evaluating SpeechLM's usefulness as an AI assistant in real-time interaction is crucial. The NTPP model \cite{ntpp} suggests using reflective pauses and interruptions to assess usefulness, where reflective pause evaluates the SpeechLM's ability to remain silent while the user speaks, and interruption measures the SpeechLM's capability to cease speaking when interrupted.

\textbf{Downstream Evaluation.} Downstream evaluation refers to evaluating the ability of SpeechLMs to perform specific tasks, such as ASR, TTS, Speaker Identification, etc. The evaluation can be performed on pre-trained models by adding few-shot example(s) at the start of the prompt or on the instruction-tuned models by directly instructing them to do so. Several benchmarks compile a broad array of downstream tasks to provide a comprehensive assessment of SpeechLMs. SUPERB \cite{yang2021superb} includes a variety of speech understanding tasks. SD-Eval \cite{sd-eval_benchmark} assesses the paralinguistic understanding of SpeechLMs using the emotion, age, environment, and age classification tasks. SALMON \cite{salmon_benchmark} tests the SpeechLMs' ability to generate speech with consistent paralinguistic and environmental characteristics. Voicebench \cite{VoiceBench} evaluates SpeechLM general capabilities. Dynamic-SUPERB \cite{dynamicsuperb_benchmark}, MMAU \cite{mmau_benchmark}, AirBench \cite{airbench_benchmark}, and AudioBench \cite{audiobench_benchmark} extend beyond traditional speech tasks to include sound and/or music-related challenges as well. 

While these benchmarks offer comprehensive coverage of audio-related tasks, they primarily require models to respond in text, which creates a barrier to end-to-end speech interaction evaluation. To address this limitation, VoxEval \cite{cui2025voxeval} focuses on benchmarking the knowledge understanding capabilities of SpeechLMs, providing question-answer pairs in comprehensive subjects in speech format, along with an evaluation pipeline tailored for speech output. Additionally, it presents questions under various input audio conditions for evaluating the model's robustness and pioneers the spoken math reasoning evaluation.

\subsection{Human (Subjective) Evaluation.}
Human evaluation plays a crucial role in assessing the performance of SpeechLMs, as ultimately, speech is designed to be heard and perceived by humans. This type of evaluation relies on human judgment to assess the quality of the outputs generated by SpeechLMs. Below, we outline several commonly used human evaluation methods.

\textbf{Mean Opinion Score.} Mean opinion score (MOS) is a widely used metric in the field of speech evaluation that quantifies the perceived quality of speech output as judged by human listeners. Typically, a group of evaluators listens to a series of audio samples generated by the SpeechLM and rates each sample on a predefined scale, often from 1 (poor quality) to 5 (excellent quality).

MOS is calculated by averaging the scores given by all evaluators for each audio sample, providing a single score that reflects the overall quality as perceived by humans. Variations of MOS focus on different aspects of speech quality, including MMOS, PMOS, and SMOS \cite{prosodyawareSLM,speechgpt-gen}. They evaluate the aspects of naturalness, prosody, and timbre similarity of the given speech, respectively.

Typically, evaluating naturalness or timbre similarity involves collecting human opinions. However, this process can be complicated due to the challenges of recruiting participants and gathering their evaluations. As a result, researchers often turn to machine-based evaluations. They commonly employ neural network models specifically trained for these tasks. For instance, a naturalness prediction model \cite{nisqanaturalness} can assess the naturalness of generated outputs, while a speaker identification model can evaluate timbre similarity.

\section{Challenges and Future Directions}
\label{sec:challenges}
While SpeechLMs have demonstrated impressive abilities, the research in this area is still under explored. In this section, we survey challenges, unsolved questions, and possible directions for future research in the study of SpeechLMs.

\subsection{Understanding Different Component Choices}
Current research on SpeechLMs encompasses key components such as speech tokenizers, language models, and vocoders, each offering a diverse range of options. While some studies have compared various component choices—primarily focusing on speech tokenizers—the comparisons tend to be limited in scope and depth \cite{OnGenerative,audiopalm}. Consequently, there remains a significant gap in understanding the advantages and disadvantages of different component selections. Therefore, studies aimed at comprehensively comparing these choices are essential. Such an investigation would yield valuable insights and serve as a guide for selecting more efficient components when developing SpeechLMs.


\subsection{End-to-End Training} 
Although SpeechLMs can generate speech directly without relying on text signals, some studies train the three components separately. This separate optimization may hinder the model's overall potential. Consequently, it would be worthwhile to investigate whether training can be conducted in an end-to-end manner, allowing gradients to be back-propagated from the vocoder's output to the tokenizer's input. By exploring this fully end-to-end approach, we could potentially enable SpeechLMs to produce more coherent, contextually relevant, and high-fidelity speech outputs.

\subsection{Real-Time Speech Generation}
Enabling real-time speech generation is crucial in SpeechLM as it fosters a more interactive way of engaging with humans. However, the most adopted approaches described in section \ref{sec:components} still result in noticeable delays between input and output speech generation. This delay occurs because a typical vocoder must wait for the entire sequence of output tokens to be generated by the language model before functioning, making it the most time-consuming process in the inference pipeline. One potential solution to improve latency is to develop a streamable pipeline, allowing the speech input and output to be processed and generated in chunks. Another option could involve the SpeechLM autonomously generating audio samples in waveform. Overall, this area of real-time speech generation remains underexplored and requires further investigation.

\subsection{Safety Risks in SpeechLMs}
\label{chall:safetyrisks}
Safety is a highly significant subject in the field of Machine Learning, particularly when it comes to large-scale generative AI models. While there has been extensive research on safety concerns in TextLMs, the safety issues in SpeechLMs have not been thoroughly investigated. The safety challenges in SpeechLMs present both similarities and unique aspects compared to TextLMs, as highlighted in OpenAI's recent report on the safety issues of GPT-4o's voice model \cite{gpt4osystemcard}. Therefore, it is crucial for future research to explore safety vulnerabilities in SpeechLMs and develop safer SpeechLMs.

Primary concerns for the safety issues in SpeechLMs include but are not limited to \textit{toxicity} and \textit{privacy}. \textit{Toxicity} refers to the harmful nature of the content generated by SpeechLMs. For instance, these models might produce semantically dangerous content, such as instructions for making explosives. Additionally, they could generate acoustically inappropriate content, like erotic speech \cite{gpt4osystemcard}, which presents a unique challenge. \textit{Privacy} involves the risk of revealing personal information from the speech input after it has been processed by a SpeechLM. For example, the model might infer the speaker's identity based on the semantic content or acoustic features of the input. Even more concerning is the potential for the model to make biased inferences about the speaker, such as their ethnicity or religious beliefs, based on insufficient (e.g., acoustic) information \cite{gpt4osystemcard}.


\subsection{Performance on Rare Languages}
SpeechLMs directly model speech data, which allows them to more effectively handle ``low-resource" languages compared to TextLMs. ``Low-resource" languages are those that lack extensive textual data, making it challenging for TextLMs to model them efficiently. In contrast, SpeechLM provides a better solution by modeling the speech data of these ``low-resource" languages, which often have more available audio data than text \cite{OnGenerative}. Therefore, future research could focus on training SpeechLMs in ``low-resource" languages or dialects to expand their capabilities.

\section{Conclusions}
This survey provides a comprehensive overview of recent advancements in Speech Language Models (SpeechLMs). We begin by addressing the limitations of the naive framework that combines Automatic Speech Recognition (ASR), Large Language Models (LLMs), and Text-to-Speech (TTS) systems for voice interactions. Next, we highlight the key advantages offered by SpeechLMs.
Following this, we explore the architectures of SpeechLMs, detailing the components involved and their training recipes. We also discuss their capabilities in various downstream applications as well as their different evaluation methods.
Finally, we identify the major challenges in developing SpeechLMs and outline potential directions for future research.
We hope this survey will illuminate the field and assist the research community in creating more powerful Speech Language Models.

\bibliographystyle{IEEEtran}
\bibliography{references}





 




\vfill

\end{document}